\documentclass[10pt,twocolumn,letterpaper]{article}

 \usepackage{cvpr}              

\usepackage{tabularx}
\usepackage{array}
\usepackage{multirow}
\usepackage{makecell}
\newcolumntype{Y}{>{\centering\arraybackslash}X}
\usepackage{placeins}
\usepackage{float}
\usepackage{capt-of}
\usepackage{dblfloatfix}
\usepackage{xcolor}
\definecolor{qualblue}{RGB}{45,105,145}
\usepackage{tabularx}
\usepackage{array}
\usepackage{multirow}
\usepackage{graphicx}

\definecolor{cvprblue}{rgb}{0.21,0.49,0.74}
\usepackage[pagebackref,breaklinks,colorlinks,allcolors=cvprblue]{hyperref}

\def\paperID{*****} 
\def\confName{CVPR}
\def\confYear{2026}

\title{LightMIS: Ultra-Lightweight Medical Image Segmentation Without a Stage-Wise Decoder}

\author{
Andrei Arhire \qquad Mihaela-Elena Breaban\\
Faculty of Computer Science\\
Alexandru Ioan Cuza University of Iasi\\
{\tt\small
\{andrei.arhire,mihaela.breaban\}@info.uaic.ro
}
\and
Radu Timofte\\
Computer Vision Lab, CAIDAS \& IFI\\
University of Wurzburg, Germany\\
{\tt\small radu.timofte@uni-wuerzburg.de}
}
\begin{document}
\maketitle
\begin{abstract}
We present LightMIS, a scalable family of ultra-lightweight
convolutional networks for 2D binary medical image
segmentation without a learned stage-wise decoder.
LightMIS aligns the outputs of a five-level encoder to a
common resolution using Scale-Aligned Projection blocks,
aggregates them once, and refines the fused representation
with an Adaptive Fusion Cascade. The cascade combines
Adaptive Kernel Fusion with the proposed Progressive
Receptive Fusion module, which uses temporary channel
expansion, complementary depthwise receptive fields, and
progressive cross-branch information transfer.

We evaluate LightMIS-T, LightMIS-S, and LightMIS using
five-fold cross-validation under a common nnU-Net v2.3.1
protocol on DRIVE, Kvasir-SEG, DSB18, BUSI, ISIC-2017,
and ISIC-2018. Full LightMIS contains \(0.131\,\mathrm{M}\)
parameters and requires \(0.575\,\mathrm{GFLOPs}\) for a
\(3\times256\times256\) input, achieving modality-macro
Dice and IoU scores of \(86.71\%\) and \(78.99\%\),
respectively. Mobile U-ViT obtains \(86.75\%\) Dice and
\(79.07\%\) IoU, so the observed differences are \(0.04\)
and \(0.08\) percentage points. Relative to Mobile U-ViT,
nnWNet, and nnU-Net, LightMIS reduces parameter count by
\(90.58\)--\(99.61\%\) and GFLOPs by
\(82.54\)--\(96.14\%\).

On an Arm Mali-G52 MC2 GPU, all LightMIS variants achieve
full GPU delegation, with median delegated latency ranging
from \(53.31\,\mathrm{ms}\) for LightMIS-T to
\(138.31\,\mathrm{ms}\) for LightMIS. These results
demonstrate a favorable accuracy--complexity trade-off and
on-device execution feasibility for the evaluated tasks.
The code is publicly available at \href{https://github.com/AndreiiArhire/LightMIS}{https://github.com/AndreiiArhire/LightMIS}.
\end{abstract}    

\section{Introduction}
\label{sec:introduction}

\begin{figure*}[t]
    \centering
    \includegraphics[width=\textwidth]
    {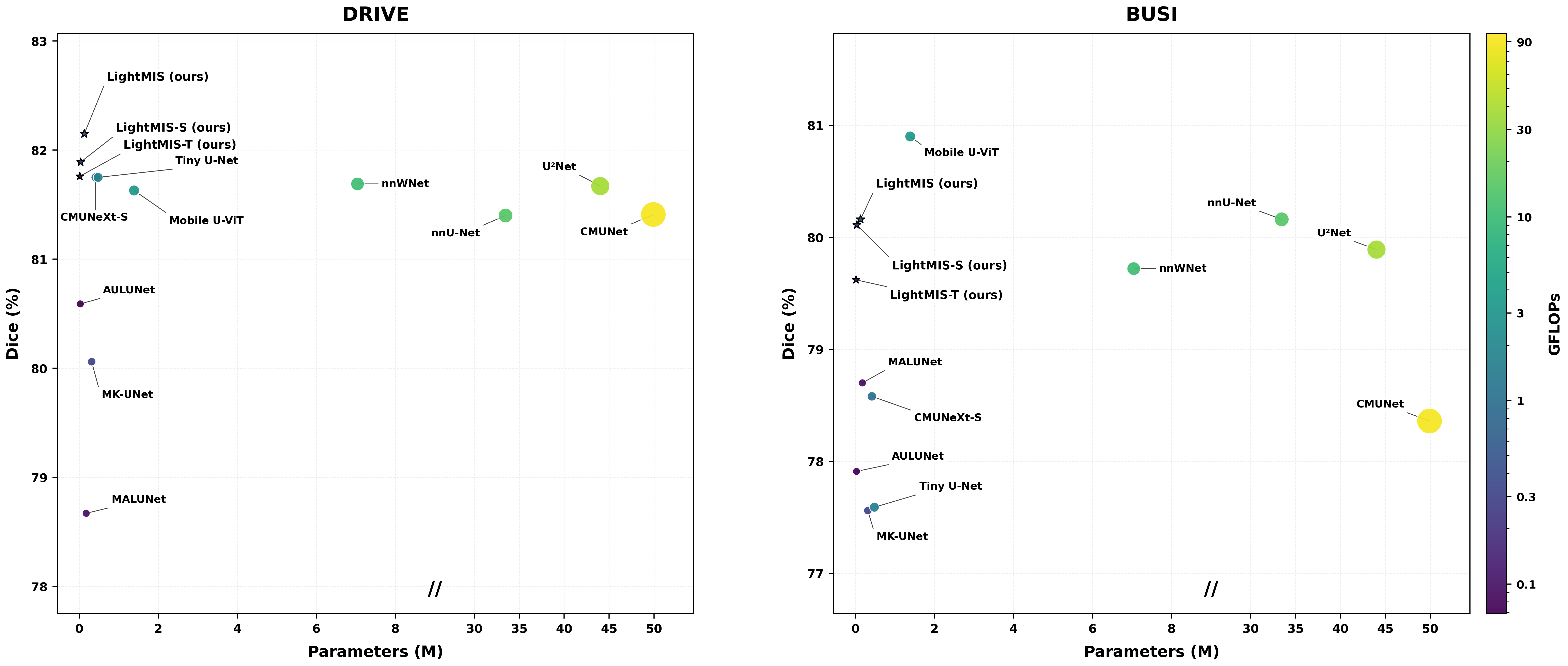}
    \caption{Accuracy--complexity trade-off on DRIVE and BUSI. The horizontal coordinate denotes trainable parameter count, the vertical coordinate denotes mean Dice over five folds, and marker color denotes GFLOPs for a \(3\times256\times256\) input. The broken horizontal axis displays compact and large models in the same panel. All models were trained from scratch under the common nnU-Net-based protocol~\cite{nnunet}.}
    \label{fig:teaser}
\end{figure*}

Accurate delineation of anatomical structures and pathological regions is essential for a wide range of computer-assisted medical imaging applications. By providing precise spatial information, medical image segmentation supports clinical workflows and reduces the workload associated with manual annotation. Although deep learning has substantially improved segmentation performance, many recent high-performing models remain computationally demanding. Their deployment may be challenging in point-of-care systems, mobile platforms, and resource-constrained clinical environments~\cite{du2026lightweight}, highlighting the practical importance of compact models that balance segmentation performance with computational efficiency. Achieving this balance is challenging because accurate medical image segmentation depends on both local details and broader spatial context. Fine details are important for locating weak boundaries and subtle changes in texture or intensity \cite{zeng2026fds}, whereas broader contextual information helps distinguish visually similar regions. Effective segmentation, therefore, requires models that preserve fine-grained features while incorporating context from larger image regions. Convolutional networks are effective at extracting local structures and boundary information while gradually incorporating broader spatial context. Although Transformer-based models capture long-range relationships through self-attention \cite{selfattention}, their use in hybrid architectures can increase computational and memory costs. This trade-off makes convolutional architectures a practical choice for resource-constrained medical image segmentation. Within this family, U-Net \cite{unet} has become one of the most widely adopted architectures for medical image segmentation. Its encoder progressively transforms high-resolution inputs into increasingly abstract feature representations, while the decoder restores spatial resolution by combining deep semantic information with fine-grained encoder features through skip connections. nnU-Net~\cite{nnunet,nnunetv2} highlighted that strong U-Net performance depends on both the architecture and the careful configuration of the overall segmentation pipeline. The framework also provides a standardized basis for evaluating architectural contributions under consistent experimental conditions. Nevertheless, conventional U-shaped networks reconstruct dense predictions through repeated upsampling and cross-scale feature fusion. This motivates the investigation of compact architectures that exploit the encoder hierarchy without relying on symmetric stage-wise decoding.

To address these challenges, we introduce LightMIS, a scalable family of ultra-lightweight medical image segmentation networks that replaces conventional stage-wise decoding with a compact head that directly aligns and aggregates the complete encoder hierarchy. Each encoder stage employs an Adaptive Fusion Cascade (AFC), which combines the Adaptive Kernel Fusion (AKF) block adopted from AULUNet~\cite{aulunet}, our proposed Progressive Receptive Fusion (PRF) module, and residual adaptive-kernel refinement. PRF enriches narrow feature representations through channel expansion, multi-kernel depthwise convolutions, and progressive cross-branch information transfer. The outputs of all encoder stages are processed using lightweight Scale-Aligned Projection (SAP) blocks, directly aggregated, and refined by a final AFC before prediction. LightMIS is integrated into nnU-Net and evaluated on six datasets under a unified training and evaluation protocol. With 0.131 M parameters and 0.575 GFLOPs, the full LightMIS achieves the second-highest Dice and IoU scores after macro-averaging across the five imaging modalities among the compared models. LightMIS-T requires only 0.017 M parameters and 0.099 GFLOPs, whereas LightMIS-S provides an intermediate operating point with 0.038 M parameters and 0.195 GFLOPs. Evaluation across overlap and boundary metrics shows that LightMIS achieves the highest mean Dice and IoU on DRIVE, whereas on DSB18 it obtains the lowest mean ASSD and HD95. A visual comparison of the proposed LightMIS variants with other models on the DRIVE and BUSI datasets is presented in Fig.~\ref{fig:teaser}. Efficiency is additionally evaluated on desktop hardware and a smartphone equipped with an Arm Mali-G52 MC2 GPU, achieving 100\% GPU operator coverage on the smartphone and supporting its applicability to resource-constrained medical image segmentation.

\paragraph{Terminology.}
In this work, \emph{without a stage-wise decoder} means that the network contains no learned sequence of resolution-recovery stages and no level-by-level encoder--decoder skip fusion. Instead, the complete encoder hierarchy is projected to a shared channel width, aligned once to a common spatial resolution, aggregated, and refined by a compact prediction head. Bilinear feature alignment and final logit resizing are retained, but no symmetric decoding path is used.

The main contributions of this work are summarized as follows:

\begin{itemize}
\item We propose Progressive Receptive Fusion (PRF), a lightweight feature-processing module that combines channel expansion, multi-kernel depthwise convolutions, and progressive cross-branch information transfer to enrich narrow feature representations.

\item We introduce the Adaptive Fusion Cascade (AFC), which sequentially combines the Adaptive Kernel Fusion (AKF) block adopted from AULUNet~\cite{aulunet}, the proposed PRF module, and residual adaptive-kernel refinement.

\item  We develop LightMIS, a scalable family of ultra-lightweight direct-aggregation segmentation networks in which Scale-Aligned Projection (SAP) blocks align the complete encoder hierarchy to a common resolution for compact aggregation, followed by AFC-based refinement and prediction. The LightMIS-T, LightMIS-S, and LightMIS configurations offer different trade-offs between segmentation accuracy and computational efficiency.

\item We integrate LightMIS into nnU-Net~\cite{nnunet,nnunetv2} and conduct a unified evaluation on DRIVE, Kvasir-SEG, DSB18, BUSI, ISIC-2017, and ISIC-2018. The evaluation covers overlap and boundary accuracy, computational complexity, memory consumption, and latency on desktop and smartphone hardware, with the full LightMIS achieving the second-highest modality-macro Dice and IoU scores among the evaluated models.

\end{itemize}

The remainder of this paper is organized as follows. Section~\ref{sec:related_work} reviews state-of-the-art and lightweight approaches to medical image segmentation. Section~\ref{sec:methodology} describes the LightMIS architecture and its main components. Section~\ref{sec:experiments} details the experimental setup and presents the quantitative and qualitative comparisons, efficiency analysis, on-device benchmarking using a smartphone GPU, and ablation studies. The limitations of this study are discussed in Section~\ref{sec:limitations}. Finally, Section~\ref{sec:conclusion} summarizes the main findings and discusses directions for future work.

\section{Related Work}
\label{sec:related_work}

The U-Net architecture~\cite{unet}, building on fully convolutional segmentation~\cite{long2015fully}, established hierarchical encoding, skip-connected feature reuse, and progressive spatial recovery as core elements of convolutional medical image segmentation. Subsequent variants modified these components without changing the overall U-shaped organization. UNet++~\cite{unetpp} introduced nested skip pathways to reduce the semantic gap between encoder and decoder features, whereas Attention U-Net~\cite{attentionunet} applied gating before skip fusion. U$^2$-Net~\cite{u2net}, originally proposed for salient object detection, uses residual U-blocks to extract intra-stage multi-scale representations. MultiResUNet~\cite{multiresunet} combines MultiRes blocks with residual paths, ResUNet++~\cite{resunetpp} integrates residual blocks with squeeze-and-excitation~\cite{senet}, atrous spatial pyramid pooling~\cite{deeplab}, and attention, and SFS-Net~\cite{sfsnet} enhances shallow representations to improve fine-grained spatial information.

Convolutional alternatives have also been developed to strengthen contextual modeling, multi-scale feature representation, and feature fusion without relying on Transformer-style self-attention~\cite{fan2020pranet,kim2021uacanet,zeng2026fds,bcunet,gfanet}. CMU-Net~\cite{cmunet} combines hybrid convolutions derived from ConvMixer~\cite{convmixer} with multi-scale attention gates, whereas CMUNeXt~\cite{cmunext} uses large-kernel depthwise convolutions, inverted bottlenecks, and skip fusion. These methods strengthen feature representation, but their hierarchical features are still integrated mainly through stage-wise decoder reconstruction~\cite{rahman2023medical,rahman2024emcad}.

Transformer-based segmentation networks use self-attention to capture dependencies between distant image regions~\cite{valanarasu2021medical, h2former, convformer, duat}. SwinUNet~\cite{swinunet} extends the U-shaped design using hierarchical Transformer blocks, whereas TransUNet~\cite{transunet} and TransAttUNet~\cite{transattunet} combine convolutional feature extraction with Transformer-based contextual modeling. Mobile U-ViT~\cite{mobileuvit} integrates large-kernel convolutional processing with a lightweight Transformer bottleneck, whereas CVPixUNet~\cite{cvpixunet} combines pixel-level Transformer modeling with adaptive convolutional processing for retinal vessel segmentation. Unlike designs that limit Transformer processing to selected stages, nnWNet~\cite{nnwnet} uses parallel convolutional and Transformer branches to support continuous transmission and multi-scale fusion of local and global features. More recently, U-Mamba~\cite{umamba} and VM-UNet~\cite{vmunet} have employed state-space modeling to capture long-range dependencies with computational complexity that scales linearly with sequence length.

Lightweight medical image segmentation networks reduce model complexity through efficient convolutional or MLP operators~\cite{liu2024rolling}, compact attention mechanisms, and resource-efficient feature-processing strategies. UNeXt~\cite{unext} combines convolutional processing with tokenized MLP blocks, whereas MALUNet~\cite{malunet} introduces dilated gated attention, external attention, and channel and spatial attention bridges. DSNET~\cite{dsnet} emphasizes detailed structures through detail-enhanced separable difference convolution and dynamic gate attention, whereas LEU-Net~\cite{leunet} incorporates multi-scale and edge-aware processing to improve boundary representation. More recently, WIDENet~\cite{widenet} combines wavelet-based multi-resolution decomposition with Transformer-based contextual encoding and attentive feature refinement to preserve structural and boundary information. These approaches illustrate complementary strategies for reducing computational requirements while maintaining effective feature representation.

Recent highly compressed U-shaped architectures place greater emphasis on preserving multi-scale and spatial information within compact model configurations. EGE-UNet~\cite{egeunet} employs grouped Hadamard-product attention together with a mask-assisted aggregation bridge, whereas LB-UNet~\cite{lbunet} combines grouped attention with auxiliary region and boundary prediction. TinyU-Net~\cite{tinyunet} improves compact feature representation through cascaded multi-receptive-field processing, MK-UNet~\cite{mkunet} introduces multi-kernel depthwise convolution within an efficient U-shaped architecture, and AULUNet~\cite{aulunet} adaptively combines local and dilated depthwise responses. UltraSeg-130K~\cite{ultraseg} combines constrained dilated convolutions with lightweight attention-guided cross-layer fusion for CPU-efficient polyp segmentation, whereas DHR-Net~\cite{dhrnet} combines dynamic staged depthwise convolution, hybrid attention, and ResECA-based cross-channel enhancement. Together, these methods demonstrate that multi-scale processing, attention, and boundary-aware feature refinement can substantially reduce model size while preserving segmentation performance. Nevertheless, they generally adopt U-shaped encoder--decoder structures that progressively integrate hierarchical features through decoding stages or stage-wise encoder--decoder fusion.

\begin{figure*}[t]
    \centering
    \includegraphics[width=\textwidth]
    {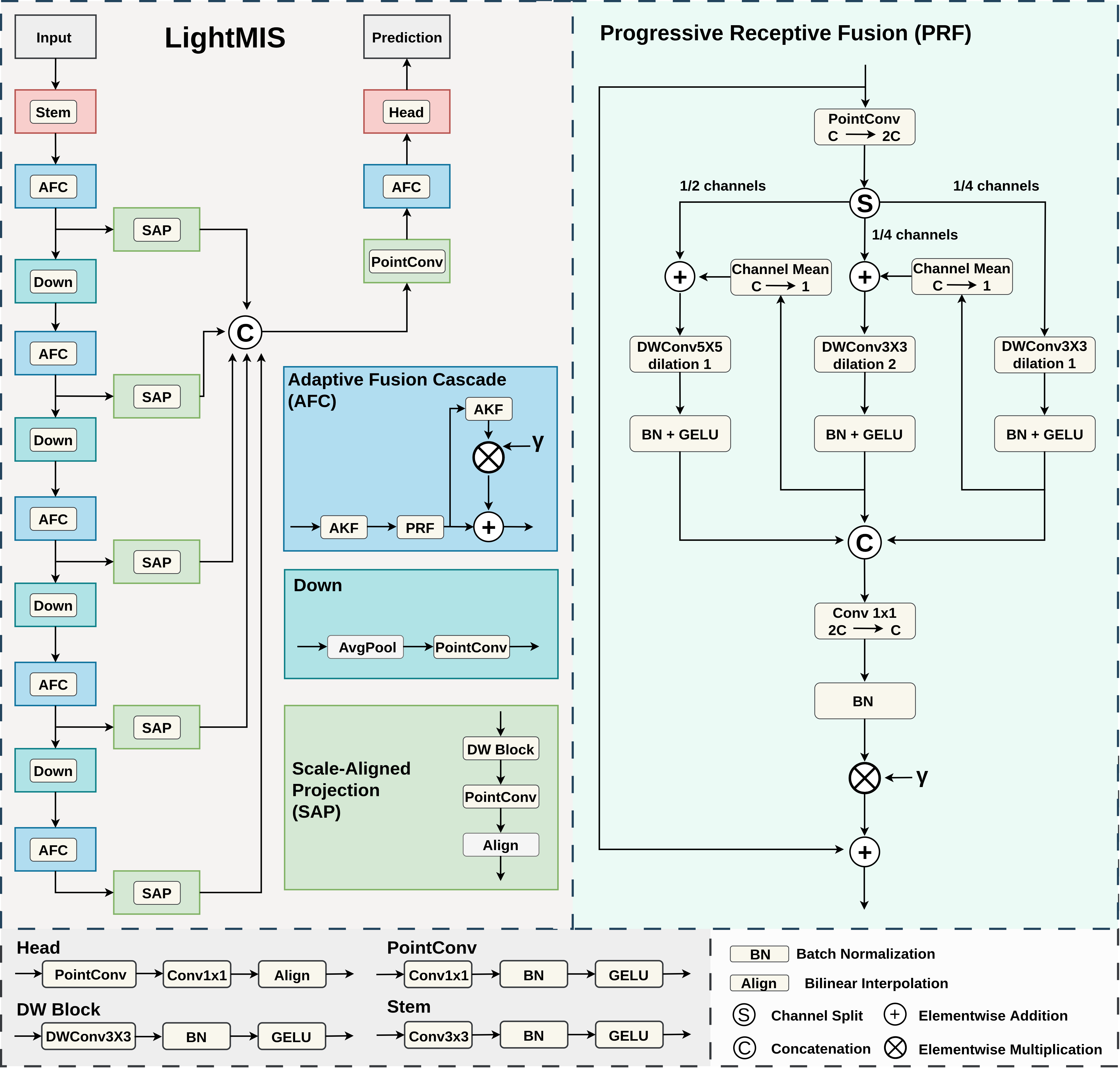}
    \caption{Overview of the proposed LightMIS architecture, highlighting the Adaptive Fusion Cascade (AFC) module and the proposed Progressive Receptive Fusion (PRF) block. The AKF block is adapted from AULUNet~\cite{aulunet}.}
    \label{fig:architecture}
\end{figure*}

Decoder-free medical image segmentation has also been explored as an alternative to conventional U-shaped reconstruction. FSE-Net~\cite{fsenet} replaces the decoder with multi-head fusion of low- and high-level features for retinal vessel segmentation, whereas CBL-Net~\cite{cblnet} recovers spatial resolution through an attention-based global feature recovery module. FAN-Net~\cite{fannet} further removes both conventional pooling and decoding by combining patch-merge attention with multi-path feature aggregation. However, their evaluation remains largely focused on retinal vessel segmentation.

We introduce LightMIS, an ultra-lightweight, scalable direct-aggregation architecture whose PRF module enriches narrow encoder features by progressively propagating channel-averaged spatial responses between heterogeneous depthwise branches with complementary receptive fields. SAP blocks align all five encoder outputs to a common resolution for direct aggregation, after which AFC refines the fused hierarchy before prediction. The same architecture is evaluated without task-specific modifications under a unified nnU-Net protocol across six datasets and five imaging settings, whereas its practical inference efficiency is assessed on both laptop and smartphone GPUs.

\section{Methodology}
\label{sec:methodology}

\subsection{LightMIS Architecture}
\label{sec:lightmis_architecture}

The overall architecture of LightMIS is illustrated in Fig.~\ref{fig:architecture}, along with the internal structure of its main feature-processing and aggregation components. Given an input image $\mathbf{X}\in\mathbb{R}^{B\times C_{\mathrm{in}}\times H\times W}$, the network extracts a five-level feature hierarchy through a compact convolutional encoder. The resulting multi-scale representations are aligned through Scale-Aligned Projection (SAP) blocks and directly integrated by a compact aggregation head. Let $\mathbf{E}_i$ denote the output of encoder level $i$, where $i\in\{0,\ldots,4\}$, with spatial dimensions progressively reduced as $H_i=H/2^i$ and $W_i=W/2^i$ and channel widths $C_i$ determined by the selected LightMIS configuration, as defined in Eq.~\ref{eq:encoder_block}.

\begin{equation}
\label{eq:encoder_block}
\mathbf{E}_i
\in
\mathbb{R}^{B\times C_i\times H_i\times W_i}
\end{equation}

In the following, $\operatorname{PWConv}(\cdot)$ denotes a $1\times1$ pointwise convolution followed by batch normalization and GELU activation. The notation $\operatorname{DW}_{k\times k, d}$ denotes a depthwise convolution with kernel size $k\times k$ and dilation rate $d$, whereas $\operatorname{DWBlock}_{k\times k,d}$ includes batch normalization and GELU activation. A $3\times3$ convolution followed by batch normalization and GELU first maps the input image to the initial encoder representation $\mathbf{Z}_0$. GELU is used for feature transformations throughout the network, whereas sigmoid is used for adaptive gating within AKF. Each encoder level applies the proposed Adaptive Fusion Cascade (AFC), whose first component is an Adaptive Kernel Fusion (AKF) block adapted from AULUNet~\cite{aulunet} by replacing its final ReLU with GELU, as defined in Eq.~\ref{eq:akf}.

\begin{equation}
\label{eq:akf}
\begin{aligned}
\mathbf{L}
&=
\operatorname{DW}_{3\times3}(\mathbf{U}),\\
\mathbf{D}
&=
\operatorname{DW}_{3\times3,d=2}(\mathbf{U}),\\
\boldsymbol{\alpha}
&=
\sigma\!\left(
\operatorname{Conv}_{1\times1}
\left(
\operatorname{GAP}(\mathbf{U})
\right)
\right),\\
\operatorname{AKF}(\mathbf{U})
&=
\operatorname{PWConv}\!\left(
\boldsymbol{\alpha}\odot\mathbf{L}
+
(1-\boldsymbol{\alpha})\odot\mathbf{D}
\right).
\end{aligned}
\end{equation}

The AKF output is processed by the proposed Progressive Receptive Fusion (PRF) module and subsequently refined by a residual AKF block, with each ResAKF instance using its own learnable scalar \(\gamma^{\mathrm{A}}\in\mathbb{R}\), initialized to 0.05. The complete AFC sequence is defined in Eq.~\ref{eq:encoder}.

\begin{equation}
\label{eq:encoder}
\begin{aligned}
\operatorname{ResAKF}(\mathbf{U})
&=
\mathbf{U}
+
\gamma^{\mathrm{A}}
\operatorname{AKF}(\mathbf{U}),\\
\operatorname{AFC}(\mathbf{U})
&=
\operatorname{ResAKF}\!\left(
\operatorname{PRF}\!\left(
\operatorname{AKF}(\mathbf{U})
\right)
\right),\\
\mathbf{E}_i
&=
\operatorname{AFC}(\mathbf{Z}_i)
\end{aligned}
\end{equation}

At each of the first four transitions between consecutive encoder levels, the feature map is reduced using $2\times2$ average pooling with stride $2$ and projected to the channel width of the next level for $i\in\{0,\ldots,3\}$, as defined in Eq.~\ref{eq:downsampling}.

\begin{equation}
\label{eq:downsampling}
\mathbf{Z}_{i+1}
=
\operatorname{PWConv}\left(
\operatorname{AvgPool}_{2\times2}(\mathbf{E}_i)
\right)
\end{equation}

The encoder outputs $\{\mathbf{E}_i\}_{i=0}^{4}$ form the multi-scale feature hierarchy passed directly to the compact aggregation head. Each encoder output is spatially processed, projected to the shared aggregation width $F$, and resized to the spatial dimensions of $\mathbf{E}_1$ using bilinear interpolation with \texttt{align\_corners=False} through a Scale-Aligned Projection (SAP) block, as defined in Eq.~\ref{eq:sap}.

\begin{equation}
\label{eq:sap}
\begin{gathered}
\bar{\mathbf{E}}_i
=
\operatorname{Align}\left(
\operatorname{PWConv}_i\left(
\operatorname{DWBlock}_{3\times3}(\mathbf{E}_i)
\right)
\right),\\[-2pt]
i\in\{0,\ldots,4\}.
\end{gathered}
\end{equation}
The five aligned representations are concatenated to form a $5F$-channel tensor and projected back to $F$ channels, after which a final AFC performs additional feature refinement, as formulated in Eq.~\ref{eq:aggregation}.

\begin{equation}
\label{eq:aggregation}
\begin{aligned}
\mathbf{M}
&=
\operatorname{PWConv}\left(
\operatorname{Concat}
\left[
\bar{\mathbf{E}}_0,\ldots,\bar{\mathbf{E}}_4
\right]
\right),\\
\mathbf{Q}
&=
\operatorname{AFC}(\mathbf{M})
\end{aligned}
\end{equation}

The refined representation is passed through a lightweight prediction head consisting of PWConv followed by a $1\times1$ output convolution, and the resulting logits are bilinearly resized to the input resolution, as defined in Eq.~\ref{eq:prediction}.

\begin{equation}
\label{eq:prediction}
\widehat{\mathbf{Y}}
=
\operatorname{Align}_{H\times W}\left(
\operatorname{Conv}_{1\times1}\left(
\operatorname{PWConv}(\mathbf{Q})
\right)
\right)
\end{equation}

LightMIS therefore avoids symmetric multi-stage decoding and instead performs direct aggregation of the complete encoder hierarchy followed by compact feature refinement and prediction. To support different resource budgets, LightMIS-T, LightMIS-S, and LightMIS preserve the same architecture while varying only their encoder channels and aggregation width, as specified in Eq.~\ref{eq:channel_schedules}. 

\begin{equation}
\label{eq:channel_schedules}
\begin{aligned}
\text{LightMIS-T}:&\ (3,6,10,16,24), \quad F=10,\\
\text{LightMIS-S}:&\ (5,10,16,24,40), \quad F=16,\\
\text{LightMIS}:&\ (10,20,32,48,80), \quad F=32
\end{aligned}
\end{equation}

The three configurations contain 0.017 M, 0.038 M, and 0.131 M parameters and require 0.099, 0.195, and 0.575 GFLOPs, respectively.

\subsection{Progressive Receptive Fusion}
\label{sec:prf}

The narrow channel configurations of LightMIS limit the representational capacity available at each encoder level. To enrich these compact representations without increasing the overall network width, we introduce Progressive Receptive Fusion (PRF). The proposed module combines temporary channel expansion, complementary depthwise receptive fields, and progressive information transfer between spatial paths.

\begin{table*}[!b]
\caption{Comparison of six medical image segmentation datasets used to evaluate our method.}
\label{tab:datasets}

\centering
\footnotesize
\renewcommand{\arraystretch}{1.1}
\setlength{\tabcolsep}{5pt}

\begin{tabular*}{\textwidth}{@{\extracolsep{\fill}}c|cccc@{}}
\hline
\textbf{Dataset}
& \textbf{Modality}
& \textbf{Imaging Region}
& \textbf{Segmentation Object}
& \textbf{No. of images} \\
\hline

DRIVE \cite{drive}
& Fundus
& Retina
& Retinal Vessel
& 40 \\

BUSI \cite{busi}
& Ultrasound
& Breast
& Breast Lesion
& 647 \\

ISIC-2017 \cite{isic2017}
& Dermoscopy
& Skin
& Lesion
& 2750 \\

ISIC-2018 \cite{isic2018}
& Dermoscopy
& Skin
& Lesion
& 3694 \\

Kvasir-SEG \cite{kvasirseg}
& Colonoscopy
& Gastrointestinal Tract
& Polyp
& 1000 \\

DSB18 \cite{dsb18}
& Microscopy
& Cell Nuclei
& Nuclei
& 670 \\

\hline
\end{tabular*}

\end{table*}

\begin{table*}[!t]
\caption{Dataset construction and evaluation protocol. All exclusions and combinations of official partitions are reported explicitly.}
\label{tab:dataset_protocol}

\centering
\footnotesize
\setlength{\tabcolsep}{1.2pt}
\renewcommand{\tabularxcolumn}[1]{m{#1}}

\begin{tabularx}{\linewidth}{
@{}
>{\raggedright\arraybackslash}m{0.105\linewidth}
|
>{\raggedright\arraybackslash}m{0.165\linewidth}
>{\centering\arraybackslash}m{0.095\linewidth}
>{\raggedright\arraybackslash}m{0.130\linewidth}
>{\centering\arraybackslash}m{0.145\linewidth}
>{\centering\arraybackslash}m{0.095\linewidth}
>{\centering\arraybackslash}m{0.105\linewidth}
Y
@{}
}
\hline

Dataset
& \shortstack{Source\\partitions} & \shortstack{Included\\cases} & Exclusions & \shortstack{Fold\\unit} & \shortstack{Input\\channels} & Resampling & \shortstack{Patch /\\batch} \\

\hline

DRIVE~\cite{drive} & Official training and test sets & 40 & None & Image & 3 & Native & $640\times640$ / 2 \\

BUSI~\cite{busi} & Benign and malignant subsets & 647 & 133 normal images excluded & Image & 3 & $256\times256$ & $256\times256$ / 32 \\

ISIC-2017~\cite{isic2017} & Official train, validation, and test sets & 2750 & None & Image & 3 & $256\times256$ & $256\times256$ / 49 \\

ISIC-2018~\cite{isic2018} & Official train, validation, and test sets & 3694 & None & Image & 3 & $256\times256$ & $256\times256$ / 49 \\

Kvasir-SEG~\cite{kvasirseg} & Official set & 1000 & None & Image & 3 & $256\times256$ & $256\times256$ / 49 \\

DSB18~\cite{dsb18} & Official labeled training set & 670 & None
& \makebox[\linewidth][c]{Image}\newline
  \makebox[\linewidth][c]{(instance-mask union)}
& 3 & $256\times256$ & $256\times256$ / 34 \\
\hline
\end{tabularx}

\end{table*}

Given an input feature map $\mathbf{U}$ with $C$ channels, PRF first expands its width to $2 \times C$ using PWConv. Channel expansion is used in several recent lightweight segmentation architectures~\cite{malunet,cmunext,mkunet,mobileuvit}, motivating its integration into PRF. The expanded representation is divided into three groups, denoted by \(\mathbf{A}_0\), \(\mathbf{B}_0\), and \(\mathbf{C}_0\), whose channel dimensions \(C_A\), \(C_B\), and \(C_C\) follow the exact integer split rule, as denoted in Eq.~\ref{eq:prf_channel_split}.

\begin{equation}
\label{eq:prf_channel_split}
C_A=C_B=\left\lfloor\frac{C}{2}\right\rfloor,
\qquad
C_C=2C-C_A-C_B.
\end{equation}

Building on the effectiveness of depthwise convolutions with varying kernel sizes or dilation rates in recent lightweight architectures~\cite{mkunet,aulunet, malunet, cmunext, egeunet, ultraseg}, the three groups are processed using different spatial operators: a depthwise $3\times3$ convolution for the first group, a dilated depthwise $3\times3$ convolution with dilation rate $2$ for the second, and a depthwise $5\times5$ convolution for the third. Each spatial operation is followed by batch normalization and GELU activation.

PRF further introduces progressive information transfer between these receptive-field paths. Although the Cascade Multi-Receptive Fields (CMRF) block in TinyU-Net~\cite{tinyunet} has proven effective for lightweight multi-scale feature processing, PRF differs by propagating a channel-averaged spatial response from each processed branch to the subsequent branch before convolution. The progressive cross-branch interaction is formulated in Eq.~\ref{eq:prf_progressive}. The channel-wise average $\operatorname{Mean}_c(\mathbf{A})$ is added to $\mathbf{B}_0$ before the dilated depthwise convolution, whereas $\operatorname{Mean}_c(\mathbf{B})$ is similarly propagated to $\mathbf{C}_0$ before the depthwise $5\times5$ convolution.

\begin{equation}
\label{eq:prf_progressive}
\begin{aligned}
\mathbf{A}
&=
\operatorname{DWBlock}_{3\times3}(\mathbf{A}_0),\\
\mathbf{B}
&=
\operatorname{DWBlock}_{3\times3,d=2}\left(
\mathbf{B}_0+\operatorname{Mean}_c(\mathbf{A})
\right),\\
\mathbf{C}
&=
\operatorname{DWBlock}_{5\times5}\left(
\mathbf{C}_0+\operatorname{Mean}_c(\mathbf{B})
\right).
\end{aligned}
\end{equation}

Channel-wise averaging compresses the preceding branch into a single spatial map that is broadcast across the channels of the next branch. This provides a representation independent of the number of channels for information transfer without requiring matching channel dimensions between the two branches. The local response progressively propagates through the dilated and $5\times5$ paths, enabling cross-branch interaction before fusion.

The three branch outputs $\mathbf{A}$, $\mathbf{B}$, and $\mathbf{C}$ are concatenated and projected back to the original channel width through a $1\times1$ convolution followed by batch normalization, producing the fused representation $\mathbf{R}$. The final PRF output combines $\mathbf{R}$ with the input $\mathbf{U}$ through a learnable scalar factor $\gamma^{\mathrm{P}}$, initialized to 0.05 separately for each PRF instance, as formulated in Eq.~\ref{eq:prf_output}.

\begin{equation}
\label{eq:prf_output}
\begin{aligned}
&\mathbf{R}
=
\operatorname{BN}\left(
\operatorname{Conv}_{1\times1}\left(
\operatorname{Concat}
[\mathbf{A},\mathbf{B},\mathbf{C}]
\right)
\right),\\
&\operatorname{PRF}(\mathbf{U})
=
\mathbf{U}+\gamma^{\mathrm{P}}\mathbf{R}.
\end{aligned}
\end{equation}

\section{Experiments and Results}
\label{sec:experiments}

\subsection{Datasets and Experimental Setup}
\label{sec:experimental_setup}

We evaluate LightMIS on six publicly available 2D binary medical image segmentation datasets covering five imaging settings to assess its performance across different modalities, anatomical regions, and segmentation targets, as summarized in Table~\ref{tab:datasets}. DRIVE~\cite{drive} addresses retinal vessel segmentation in fundus images, where the target structures are thin and highly branched. Kvasir-SEG~\cite{kvasirseg} focuses on gastrointestinal polyp segmentation in endoscopic images, with substantial variation in polyp size, shape, and surrounding tissue. DSB18~\cite{dsb18} considers cell nuclei segmentation in microscopy images acquired under different imaging conditions. BUSI~\cite{busi} evaluates breast lesion segmentation in ultrasound images, where low contrast, speckle noise, and poorly defined boundaries pose important segmentation challenges. ISIC-2017~\cite{isic2017} and ISIC-2018~\cite{isic2018} address skin lesion segmentation in dermoscopic images, including variation in lesion morphology, texture, and boundary characteristics.  Both ISIC-2017 and ISIC-2018 were included because of their widespread use as benchmarks in lightweight medical image segmentation~\cite{dsnet, leunet, widenet, aulunet, mkunet, egeunet, malunet, tinyunet}.

Table~\ref{tab:dataset_protocol} summarizes dataset construction, conversion performed prior to nnU-Net preprocessing, and evaluation settings. BUSI contains 647 lesion images, comprising 437 benign and 210 malignant cases, after excluding the 133 normal images. For DSB18, the instance masks associated with each image were combined by pixel-wise logical union to obtain a single binary semantic ground-truth mask.

All evaluated architectures are integrated into a common nnU-Net v2.3.1 pipeline using the 2D configuration. Following the preprocessing protocol used in nnWNet~\cite{nnwnet}, DRIVE images are kept at their original resolution, whereas images from the remaining datasets are resized to $256\times256$ before being processed by the nnU-Net pipeline. All inputs were represented using three channels, with grayscale images replicated across channels. Lanczos interpolation was used for images and nearest-neighbor interpolation for masks. The nnU-Net plans applied z-score normalization to each input channel without mask-based normalization. After this initial spatial standardization, dataset-specific preprocessing, patch size, and batch size are determined by the nnU-Net planning procedure. Identical data partitions, preprocessing configurations, and training settings are used across all compared architectures. No subject-level grouping or category-based stratification was applied. Following the unified evaluation strategy employed in nnWNet, all models are trained from scratch for 200 epochs using five-fold cross-validation at the image level. Deep supervision is disabled for all experiments, and no post-processing is applied to the predicted masks. Final binary masks are obtained through a pixel-wise argmax over the two softmax probabilities. For all models, only the final training checkpoint was used for inference and evaluation. Optimization is performed using SGD with Nesterov momentum of $0.99$, an initial learning rate of $10^{-2}$, and a weight decay of $3\times10^{-5}$. The learning rate follows the polynomial decay schedule used by nnU-Net. The reported mean and standard deviation for each metric are computed across the five cross-validation folds. All experiments are implemented in PyTorch v2.11.0 on the same workstation.

\begin{table*}[!b]
\caption{Comparison with representative methods on Kvasir-SEG and DRIVE. All models were trained from scratch in nnU-Net~\cite{nnunet}, using the official, publicly available implementations for the compared methods. Values are mean$\pm$standard deviation over five folds.}

\label{tab:results_kvasir_drive}
\centering
\scriptsize
\setlength{\tabcolsep}{2pt}
\renewcommand{\arraystretch}{1.05}

\resizebox{\textwidth}{!}{
\begin{tabular}{@{}ll|c|cccc|cccc@{}}
\hline

\multirow{2}{*}{Arch.} & \multirow{2}{*}{Model} 
& Params $\downarrow$
& \multicolumn{4}{c|}{Kvasir-SEG\cite{kvasirseg}} 
& \multicolumn{4}{c}{DRIVE\cite{drive}} \\

 &  & (M) 
& Dice$\uparrow$ & IoU$\uparrow$ & HD95$\downarrow$ & ASSD$\downarrow$
& Dice$\uparrow$ & IoU$\uparrow$ & HD95$\downarrow$ & ASSD$\downarrow$ \\
\hline

\multirow{16}{*}{Conv} 

& U$^{2}$Net \cite{u2net} & 44.024
& 89.60$\pm$1.57 & 84.23$\pm$1.76 & \textbf{18.43$\pm$2.20} & 5.54$\pm$0.99
& 81.67$\pm$1.24 & 69.16$\pm$1.58 & 4.67$\pm$1.47 & 1.15$\pm$0.18 \\

& nnU-Net \cite{nnunet} & 33.472
& \textbf{90.02$\pm$1.82} & \textit{84.29$\pm$2.26} & 20.37$\pm$4.33 & 5.66$\pm$1.40
& 81.40$\pm$0.83 & 68.73$\pm$1.07 & 4.62$\pm$1.27 & 1.13$\pm$0.14 \\

& UNeXt \cite{unext} & 1.472
& 87.26$\pm$1.59 & 81.17$\pm$1.65 & 22.03$\pm$2.14 & 6.63$\pm$0.83
& 81.26$\pm$0.92 & 68.55$\pm$1.16 & 5.85$\pm$1.44 & 1.27$\pm$0.15 \\

& MALUNet \cite{malunet} & 0.178
& 85.75$\pm$2.07 & 78.61$\pm$2.27 & 23.87$\pm$2.94 & 7.64$\pm$1.39
& 78.67$\pm$1.29 & 64.98$\pm$1.56 & 10.47$\pm$1.77 & 1.90$\pm$0.25 \\

& EGE-UNet \cite{egeunet} & 0.053 
& 84.08$\pm$2.57 & 76.59$\pm$2.96 & 27.06$\pm$4.73 & 8.71$\pm$2.15
& 79.78$\pm$1.27 & 66.54$\pm$1.54 & 8.80$\pm$1.76 & 1.68$\pm$0.26 \\

& CMUNet \cite{cmunet} & 49.932 
& 89.76$\pm$1.72 & \textbf{84.31$\pm$1.98} & 19.15$\pm$2.67 & \textit{5.46$\pm$0.90}
& 81.41$\pm$1.27 & 68.79$\pm$1.61 & 5.07$\pm$1.95 & 1.22$\pm$0.26 \\

& LB-UNet \cite{lbunet} & 0.056 
& 85.49$\pm$1.58 & 78.32$\pm$1.83 & 24.72$\pm$2.49 & 7.65$\pm$1.03
& 79.70$\pm$1.33 & 66.44$\pm$1.58 & 8.86$\pm$1.73 & 1.69$\pm$0.26 \\

& CMUNeXt-S \cite{cmunext} & 0.418
& 87.54$\pm$1.56 & 81.43$\pm$1.62 & 22.61$\pm$1.70 & 6.56$\pm$0.70
& 81.75$\pm$1.03 & 69.26$\pm$1.31 & 5.10$\pm$1.46 & 1.17$\pm$0.17 \\

& TinyU-Net \cite{tinyunet} & 0.481
& 88.08$\pm$1.97 & 82.29$\pm$1.98 & 20.12$\pm$1.85 & 5.60$\pm$0.64
& 81.75$\pm$0.96 & 69.27$\pm$1.20 & 5.28$\pm$1.55 & 1.20$\pm$0.18 \\

& AULUNet \cite{aulunet} & \textit{0.029}
& 84.60$\pm$1.96 & 77.41$\pm$2.23 & 26.43$\pm$3.56 & 8.40$\pm$1.48
& 80.59$\pm$0.95 & 67.65$\pm$1.15 & 7.41$\pm$1.55 & 1.48$\pm$0.19 \\

& MK-UNet \cite{mkunet} & 0.316
& 85.72$\pm$1.55 & 78.82$\pm$1.85 & 26.86$\pm$3.20 & 7.80$\pm$1.14
& 80.06$\pm$1.45 & 66.89$\pm$1.84 & 6.26$\pm$1.90 & 1.34$\pm$0.22 \\

& UltraSeg130K \cite{ultraseg} & 0.148
& 84.24$\pm$1.64 & 76.84$\pm$1.60 & 26.44$\pm$2.04 & 8.04$\pm$0.78
& 78.72$\pm$1.67 & 65.13$\pm$1.96 & 9.95$\pm$2.37 & 1.87$\pm$0.39 \\

& DHRNet \cite{dhrnet} & 0.065 
& 86.99$\pm$1.56 & 80.25$\pm$1.63 & 22.51$\pm$1.75 & 6.87$\pm$0.90
& 78.27$\pm$1.40 & 64.45$\pm$1.68 & 11.09$\pm$1.67 & 2.01$\pm$0.25 \\

\cline{2-11}
& \textbf{LightMIS-T (ours)} & \textbf{0.017} 
& 87.52$\pm$1.30 & 80.96$\pm$1.45 & 22.66$\pm$2.61 & 6.78$\pm$0.99
& 81.76$\pm$1.00 & 69.26$\pm$1.28 &  5.91$\pm$1.48 & 1.27$\pm$0.16 \\

& \textbf{LightMIS-S (ours)} & 0.038 
& 88.47$\pm$1.96 & 82.37$\pm$2.27 & 20.19$\pm$2.78 & 5.96$\pm$1.14
& \textit{81.89$\pm$0.99} & \textit{69.48$\pm$1.25} & 5.57$\pm$1.47 & 1.23$\pm$0.15 \\

& \textbf{LightMIS (ours)} & 0.131
& \textit{89.87$\pm$1.53} & 84.25$\pm$1.70 & \textit{19.14$\pm$1.98} & \textbf{5.44$\pm$0.81}
& \textbf{82.15$\pm$0.98} & \textbf{69.82$\pm$1.29} & 5.18$\pm$1.39 & 1.19$\pm$0.17 \\

\hline

Trans 
& SwinUNet \cite{swinunet} & 27.176
& 80.15$\pm$2.84 & 72.21$\pm$2.94 & 32.00$\pm$3.58 & 10.69$\pm$1.54
& 80.30$\pm$0.80 & 67.20$\pm$1.01 & 6.79$\pm$0.93 & 1.39$\pm$0.10 \\

\hline

\multirow{4}{*}{Hybrid} 

& TransAttUNet \cite{transattunet} & 25.966
& 89.30$\pm$1.52 & 83.47$\pm$1.63 & 21.36$\pm$1.81 & 5.86$\pm$0.98
& 81.39$\pm$1.14 & 68.75$\pm$1.46 & \textbf{4.26$\pm$1.23} & \textbf{1.09$\pm$0.14} \\

& TransUNet \cite{transunet} & 91.719 
& 77.69$\pm$2.91 & 68.94$\pm$2.83 & 38.37$\pm$3.92 & 11.93$\pm$1.76
& 78.42$\pm$0.98 & 64.63$\pm$1.15 & 7.95$\pm$1.51 & 1.57$\pm$0.19 \\

& nnWNet \cite{nnwnet} & 7.044
& 89.69$\pm$1.96 & 84.25$\pm$2.19 & 19.30$\pm$3.29 & 5.54$\pm$1.29
& 81.69$\pm$0.93 & 69.14$\pm$1.24 & \textit{4.36$\pm$1.21} & \textit{1.10$\pm$0.14} \\

& Mobile U-ViT \cite{mobileuvit} & 1.390
& 89.65$\pm$1.56 & 84.12$\pm$1.83 & 19.53$\pm$2.77 & 5.70$\pm$1.07
& 81.63$\pm$0.97 & 69.10$\pm$1.20 & 5.14$\pm$1.43 & 1.18$\pm$0.16 \\

\hline

\end{tabular}
}
\end{table*}

The training objective combines cross-entropy and soft Dice losses with equal weighting, as defined in Eq.~\ref{eq:train_loss}.

\begin{equation}
    \label{eq:train_loss}
    \mathcal{L}_{\mathrm{train}}
    =
    \mathcal{L}_{\mathrm{CE}}
    +
    \mathcal{L}_{\mathrm{Dice}}
\end{equation}

For binary segmentation, the head outputs background and foreground logits. After channel-wise softmax, $p_{i,c}$ denotes the probability of class $c\in\{0,1\}$, whereas $g_i\in\{0,1\}$ denotes the ground-truth class at pixel $i$. Cross-entropy uses $p_{i,g_i}$, whereas soft Dice uses the foreground-class probability, denoted by $p_i$. Accordingly, the pixel-wise cross-entropy loss is defined in Eq.~\ref{eq:ce_loss}.

\begin{equation}
    \label{eq:ce_loss}
    \mathcal{L}_{\mathrm{CE}}
    =
    -\frac{1}{N}
    \sum_{i=1}^{N}
    \log p_{i,g_i},
\end{equation}

The soft Dice loss promotes foreground overlap between the predicted probabilities $p_i$ and the binary ground-truth labels $g_i$, with $\epsilon=10^{-5}$ included for numerical stability, as formulated in Eq.~\ref{eq:dice_loss}.

\begin{equation}
    \label{eq:dice_loss}
    \mathcal{L}_{\mathrm{Dice}}
    =
    -
    \frac{
        2\sum_{i=1}^{N} p_i g_i + \epsilon
    }{
        \sum_{i=1}^{N} p_i
        +
        \sum_{i=1}^{N} g_i
        +
        \epsilon
    }
\end{equation}

Segmentation performance is evaluated using two region-overlap metrics, the Dice similarity coefficient (Dice) and Intersection over Union (IoU), together with two boundary distance metrics, the 95th percentile Hausdorff distance (HD95) and Average Symmetric Surface Distance (ASSD). The Dice score in Eq.~\ref{eq:dice} measures segmentation similarity by comparing twice the shared area of the predicted region $\mathcal{P}$ and the reference region $\mathcal{G}$ with the sum of their individual areas.

\begin{equation}
\label{eq:dice}
\mathrm{Dice}
=
\frac{2|\mathcal{P}\cap\mathcal{G}|}
{|\mathcal{P}|+|\mathcal{G}|}
\end{equation}

The IoU in Eq.~\ref{eq:iou} measures the agreement between the predicted and reference regions based on the ratio of their intersection to their union.
\begin{equation}
    \label{eq:iou}
    \mathrm{IoU}
    =
    \frac{|\mathcal{P}\cap\mathcal{G}|}
         {|\mathcal{P}\cup\mathcal{G}|}
\end{equation}

HD95 measures the 95th percentile of the distances between the predicted and reference boundaries, reducing the influence of isolated extreme deviations. ASSD measures the average bidirectional distance between the predicted and reference boundaries, reflecting their overall spatial alignment. Thus, higher Dice and IoU values indicate better region agreement, whereas lower HD95 and ASSD values indicate better boundary localization.

HD95 and ASSD are reported in pixels and are computed on the binary masks at the evaluation resolution using unit pixel spacing. When both the prediction and ground-truth masks are empty, Dice and IoU are set to 1 and both boundary distances to 0. When only one mask is empty, Dice and IoU are set to 0, whereas HD95 and ASSD are considered undefined and excluded from their respective averages.

\subsection{Comparison Under a Unified Training Protocol}
\label{sec:sota}

\begin{table*}[!t]
\caption{Comparison with representative methods on BUSI and DSB18. All models were trained from scratch in nnU-Net~\cite{nnunet}, using the official, publicly available implementations for the compared methods. Values are mean$\pm$standard deviation over five folds.}
\label{tab:results_busi_dsb18}
\centering
\scriptsize
\setlength{\tabcolsep}{2pt}
\renewcommand{\arraystretch}{1.05}

\resizebox{\textwidth}{!}{
\begin{tabular}{@{}ll|c|cccc|cccc@{}}
\hline

\multirow{2}{*}{Arch.} & \multirow{2}{*}{Model} 
& Params $\downarrow$
& \multicolumn{4}{c|}{BUSI \cite{busi}} 
& \multicolumn{4}{c}{DSB18 \cite{dsb18}} \\

 &  & (M) 
& Dice$\uparrow$ & IoU$\uparrow$ & HD95$\downarrow$ & ASSD$\downarrow$
& Dice$\uparrow$ & IoU$\uparrow$ & HD95$\downarrow$ & ASSD$\downarrow$ \\
\hline

\multirow{16}{*}{Conv}

& U$^{2}$Net \cite{u2net} & 44.024
& 79.89$\pm$1.59 & 71.97$\pm$1.46 & 23.47$\pm$2.21 & 9.03$\pm$0.97
& 91.73$\pm$0.45 & 85.55$\pm$0.71 & 4.99$\pm$1.74 & 1.67$\pm$0.67 \\

& nnU-Net \cite{nnunet} & 33.472
& \textit{80.16$\pm$1.50} & 72.21$\pm$1.44 & 24.69$\pm$2.35 & 9.38$\pm$1.36
& 91.58$\pm$0.66 & 85.30$\pm$0.89 & 4.99$\pm$1.68 & 1.66$\pm$0.70 \\

& UNeXt \cite{unext} & 1.472
& 78.16$\pm$2.33 & 70.17$\pm$2.02 & 25.51$\pm$3.49 & 9.89$\pm$1.37
& 91.50$\pm$0.52 & 85.17$\pm$0.71 & 4.69$\pm$1.91 & 1.61$\pm$0.76 \\

& MALUNet \cite{malunet} & 0.178
& 78.70$\pm$1.87 & 70.44$\pm$1.77 & 24.43$\pm$2.42 & 9.73$\pm$1.40
& 90.81$\pm$0.86 & 84.08$\pm$1.16 & 5.31$\pm$1.87 & 1.70$\pm$0.76 \\

& EGE-UNet \cite{egeunet} & 0.053 
& 77.25$\pm$1.22 & 68.62$\pm$1.22 & 26.24$\pm$2.59 & 10.54$\pm$1.17
& 90.86$\pm$0.85 & 84.15$\pm$1.13 & 5.49$\pm$2.06 & 1.69$\pm$0.77 \\

& CMUNet \cite{cmunet} & 49.932
& 78.36$\pm$1.56 & 70.64$\pm$1.47 & 24.79$\pm$3.54 & 9.85$\pm$1.83
& \textit{91.94$\pm$0.72} & \textit{85.76$\pm$1.00} & 4.75$\pm$2.14 & 1.60$\pm$0.77 \\

& LB-UNet \cite{lbunet} & 0.056 
& 77.52$\pm$1.43 & 69.08$\pm$1.11 & 26.02$\pm$2.55 & 10.21$\pm$1.35
& 90.62$\pm$0.59 & 83.86$\pm$0.82 & 5.69$\pm$1.61 & 1.75$\pm$0.73 \\

& CMUNeXt-S \cite{cmunext} & 0.418
& 78.58$\pm$1.69 & 70.65$\pm$1.54 & 28.43$\pm$3.41 & 10.70$\pm$1.50
& 91.83$\pm$0.69 & 85.66$\pm$0.89 & 5.06$\pm$1.95 & 1.61$\pm$0.79 \\

& TinyU-Net \cite{tinyunet} & 0.481
& 77.59$\pm$2.57 & 69.83$\pm$2.41 & 26.01$\pm$4.29 & 9.85$\pm$1.55
& \textbf{92.10$\pm$0.65} & \textbf{86.02$\pm$0.88} & 4.71$\pm$2.01 & \textit{1.57$\pm$0.80} \\

& AULUNet \cite{aulunet} & \textit{0.029}
& 77.91$\pm$2.20 & 69.71$\pm$2.06 & 24.83$\pm$3.91 & 10.06$\pm$1.99
& 91.00$\pm$0.86 & 84.55$\pm$1.07 & 5.75$\pm$1.70 & 1.75$\pm$0.74 \\

& MK-UNet \cite{mkunet} & 0.316
& 77.56$\pm$0.79 & 68.89$\pm$0.87 & 29.47$\pm$1.16 & 11.01$\pm$0.30
& 91.25$\pm$0.78 & 84.72$\pm$1.07 & 4.94$\pm$1.84 & 1.63$\pm$0.75 \\

& UltraSeg130K \cite{ultraseg} & 0.148 
& 77.98$\pm$2.06 & 69.62$\pm$1.94 & 25.43$\pm$3.36 & 10.01$\pm$1.81
& 90.69$\pm$0.85 & 83.93$\pm$1.21 & 5.23$\pm$1.86 & 1.71$\pm$0.72 \\

& DHRNet \cite{dhrnet} & 0.065 
& 78.36$\pm$2.08 & 70.17$\pm$1.85 & 24.84$\pm$3.13 & 9.98$\pm$1.59
& 90.69$\pm$0.87 & 83.87$\pm$1.16 & 5.09$\pm$2.02 & 1.70$\pm$0.79 \\

\cline{2-11}

& \textbf{LightMIS-T (ours)} & \textbf{0.017} 
& 79.62$\pm$1.28 & 71.48$\pm$0.96 & 23.24$\pm$1.33 & \textit{8.70$\pm$0.77}
& 91.29$\pm$0.87 & 84.89$\pm$1.20 & 5.07$\pm$2.03 & 1.67$\pm$0.80 \\

& \textbf{LightMIS-S (ours)} & 0.038 
& 80.11$\pm$1.41 & \textit{72.31$\pm$1.03} & \textit{22.12$\pm$2.03} & \textbf{8.63$\pm$1.19}
& 91.56$\pm$0.81 & 85.25$\pm$1.15 & 5.03$\pm$2.07 & 1.61$\pm$0.79 \\

& \textbf{LightMIS (ours)} & 0.131
& \textit{80.16$\pm$1.23} & 72.28$\pm$1.25 & 23.32$\pm$1.05 & 8.97$\pm$0.24
& 91.81$\pm$0.67 & 85.62$\pm$0.93 & \textbf{4.59$\pm$1.97} & \textbf{1.53$\pm$0.72} \\

\hline

Trans 
& SwinUNet \cite{swinunet} & 27.176
& 71.45$\pm$2.07 & 61.97$\pm$1.72 & 33.12$\pm$2.69 & 13.70$\pm$1.52
& 91.19$\pm$0.70 & 84.75$\pm$0.97 & 5.48$\pm$1.98 & 1.73$\pm$0.75 \\

\hline

\multirow{4}{*}{Hybrid} 

& TransAttUNet \cite{transattunet} & 25.966
& 79.45$\pm$0.83 & 71.10$\pm$0.78 & 27.23$\pm$2.34 & 9.99$\pm$0.81
& 91.69$\pm$0.59 & 85.37$\pm$0.76 & 4.79$\pm$1.98 & 1.63$\pm$0.80 \\

& TransUNet \cite{transunet} & 91.719 
& 71.07$\pm$2.97 & 62.08$\pm$2.50 & 33.49$\pm$3.09 & 13.50$\pm$2.09
& 90.00$\pm$0.63 & 82.95$\pm$0.83 & 5.40$\pm$1.61 & 1.85$\pm$0.64 \\

& nnWNet \cite{nnwnet} & 7.044
& 79.72$\pm$1.94 & 72.09$\pm$1.66 & \textbf{22.07$\pm$1.96} & 8.75$\pm$1.03
& 91.84$\pm$0.74 & 85.66$\pm$1.02 & 4.70$\pm$1.88 & \textit{1.57$\pm$0.76} \\

& Mobile U-ViT \cite{mobileuvit} & 1.390
& \textbf{80.90$\pm$1.18} & \textbf{73.27$\pm$0.83} & 22.48$\pm$2.02 & \textit{8.70$\pm$1.04}
& 91.78$\pm$0.67 & 85.54$\pm$0.97 & \textit{4.66$\pm$2.17} & 1.59$\pm$0.77 \\

\hline

\end{tabular}
}
\end{table*}

\begin{table*}[!t]
\caption{Comparison with representative methods on ISIC-2017 and ISIC-2018. All models were trained from scratch in nnU-Net~\cite{nnunet}, using the official, publicly available implementations for the compared methods. Values are mean$\pm$standard deviation over five folds.}

\label{tab:results_isic}
\centering
\scriptsize
\setlength{\tabcolsep}{2pt}
\renewcommand{\arraystretch}{1.05}

\resizebox{\textwidth}{!}{
\begin{tabular}{@{}ll|c|cccc|cccc@{}}
\hline
\multirow{2}{*}{Arch.} & \multirow{2}{*}{Model} 
& Params $\downarrow$
& \multicolumn{4}{c|}{ISIC-2017 \cite{isic2017}} 
& \multicolumn{4}{c}{ISIC-2018\cite{isic2018}}\\

 &  & (M) 
& Dice$\uparrow$ & IoU$\uparrow$ & HD95$\downarrow$ & ASSD$\downarrow$
& Dice$\uparrow$ & IoU$\uparrow$ & HD95$\downarrow$ & ASSD$\downarrow$ \\
\hline

\multirow{16}{*}{Conv} 

& U$^{2}$Net \cite{u2net} & 44.024
& \textit{89.55$\pm$0.71} & \textit{82.86$\pm$0.91} & \textit{12.71$\pm$0.57} & \textbf{5.32$\pm$0.34}
& \textbf{90.23$\pm$0.33} & \textbf{83.94$\pm$0.37} & \textbf{12.97$\pm$0.54} & \textbf{5.40$\pm$0.23} \\

& nnU-Net \cite{nnunet} & 33.472
& 89.43$\pm$0.82 & 82.72$\pm$0.97 & 13.42$\pm$1.07 & 5.64$\pm$0.61
& \textit{90.02$\pm$0.40} & 83.63$\pm$0.47 & 13.59$\pm$0.70 & 5.64$\pm$0.31 \\

& UNeXt \cite{unext} & 1.472
& 89.20$\pm$0.70 & 82.41$\pm$0.83 & 13.16$\pm$0.69 & 5.45$\pm$0.35
& 89.68$\pm$0.29 & 83.20$\pm$0.36 & 13.65$\pm$0.45 & 5.55$\pm$0.20 \\

& MALUNet \cite{malunet} & 0.178
& 89.11$\pm$0.63 & 82.22$\pm$0.81 & 13.05$\pm$0.61 & 5.58$\pm$0.39
& 89.48$\pm$0.41 & 82.90$\pm$0.52 & 13.83$\pm$0.49 & 5.88$\pm$0.23 \\

& EGE-UNet \cite{egeunet} & 0.053 
& 88.20$\pm$0.61 & 81.20$\pm$0.84 & 13.86$\pm$0.73 & 5.82$\pm$0.37
& 89.19$\pm$0.31 & 82.49$\pm$0.39 & 14.28$\pm$0.64 & 5.93$\pm$0.25 \\

& CMUNet \cite{cmunet} & 49.932
& 88.77$\pm$0.81 & 81.96$\pm$0.94 & 13.93$\pm$1.04 & 5.84$\pm$0.59
& 89.79$\pm$0.49 & 83.37$\pm$0.55 & 13.51$\pm$0.77 & 5.60$\pm$0.40 \\

& LB-UNet \cite{lbunet} & 0.056  
& 88.94$\pm$0.67 & 82.09$\pm$0.80 & 13.15$\pm$0.62 & 5.62$\pm$0.40
& 89.46$\pm$0.52 & 82.91$\pm$0.65 & 13.96$\pm$0.70 & 5.86$\pm$0.40 \\

& CMUNeXt-S \cite{cmunext} & 0.418
& 88.79$\pm$0.76 & 81.96$\pm$0.85 & 13.90$\pm$0.53 & 5.72$\pm$0.35
& 89.35$\pm$0.37 & 82.71$\pm$0.46 & 14.89$\pm$0.76 & 5.91$\pm$0.24 \\

& TinyU-Net \cite{tinyunet} & 0.481
& 89.25$\pm$0.53 & 82.49$\pm$0.62 & 13.21$\pm$0.67 & 5.42$\pm$0.28
& 89.63$\pm$0.26 & 83.13$\pm$0.34 & 14.16$\pm$0.39 & 5.66$\pm$0.19 \\

& AULUNet \cite{aulunet} & \textit{0.029} 
& 88.85$\pm$0.81 & 82.01$\pm$0.98 & 13.24$\pm$0.79 & 5.64$\pm$0.47
& 89.67$\pm$0.41 & 83.17$\pm$0.47 & 13.36$\pm$0.36 & 5.63$\pm$0.21 \\

& MK-UNet \cite{mkunet} & 0.316
& 88.73$\pm$0.75 & 81.73$\pm$0.91 & 13.91$\pm$0.65 & 5.78$\pm$0.37
& 89.39$\pm$0.42 & 82.77$\pm$0.51 & 14.26$\pm$0.72 & 5.86$\pm$0.29 \\

& UltraSeg130K \cite{ultraseg} & 0.148
& 89.09$\pm$0.62 & 82.28$\pm$0.73 & 13.18$\pm$0.45 & 5.56$\pm$0.33
& 89.77$\pm$0.29 & 83.24$\pm$0.36 & 13.60$\pm$0.37 & 5.61$\pm$0.14 \\

& DHRNet \cite{dhrnet} & 0.065 
& 89.15$\pm$0.70 & 82.37$\pm$0.85 & 12.80$\pm$0.63 & 5.49$\pm$0.32
& 89.77$\pm$0.40 & 83.25$\pm$0.49 & 13.38$\pm$0.50 & 5.62$\pm$0.28 \\

\cline{2-11}
& \textbf{LightMIS-T (ours)} & \textbf{0.017} 
& 89.09$\pm$0.63 & 82.30$\pm$0.77 & 12.99$\pm$0.53 & 5.50$\pm$0.35
& 89.69$\pm$0.56 & 83.21$\pm$0.64 & 13.62$\pm$0.60 & 5.64$\pm$0.34 \\

& \textbf{LightMIS-S (ours)} & 0.038 
& 89.20$\pm$0.70 & 82.42$\pm$0.91 & 12.90$\pm$0.71 & 5.46$\pm$0.40
& 89.74$\pm$0.44 & 83.27$\pm$0.53 & 13.46$\pm$0.43 &  5.70$\pm$0.23 \\

& \textbf{LightMIS (ours)} & 0.131
& 89.32$\pm$0.62 & 82.60$\pm$0.76 & 12.72$\pm$0.57 & \textit{5.37$\pm$0.31}
& 89.83$\pm$0.24 & 83.39$\pm$0.24 & 13.25$\pm$0.42 & \textit{5.48$\pm$0.13} \\

\hline

Trans 
& SwinUNet \cite{swinunet} & 27.176
& 88.00$\pm$0.64 & 80.97$\pm$0.77 & 14.69$\pm$0.73 & 6.13$\pm$0.44
& 88.95$\pm$0.19 & 82.28$\pm$0.17 & 14.81$\pm$0.38 & 6.05$\pm$0.20 \\

\hline

\multirow{4}{*}{Hybrid} 

& TransAttUNet \cite{transattunet} & 25.966
& 88.93$\pm$0.52 & 82.10$\pm$0.64 & 13.92$\pm$0.38 & 5.63$\pm$0.26
& 89.41$\pm$0.20 & 82.85$\pm$0.19 & 14.50$\pm$0.58 & 5.80$\pm$0.13 \\

& TransUNet \cite{transunet} & 91.719 
& 86.56$\pm$0.60 & 79.20$\pm$0.65 & 17.04$\pm$0.56 & 6.86$\pm$0.28
& 87.35$\pm$0.31 & 80.27$\pm$0.31 & 17.55$\pm$0.57 & 7.04$\pm$0.27 \\

& nnWNet \cite{nnwnet} & 7.044
& 89.38$\pm$0.58 & 82.67$\pm$0.74 & 13.08$\pm$0.74 & 5.38$\pm$0.31
& 89.84$\pm$0.21 & 83.46$\pm$0.29 & 13.51$\pm$0.39 & 5.51$\pm$0.16 \\

& Mobile U-ViT \cite{mobileuvit} & 1.390
& \textbf{89.59$\pm$0.62} & \textbf{82.97$\pm$0.76} & \textbf{12.58$\pm$0.73} & \textbf{5.32$\pm$0.43}
& 89.99$\pm$0.26 & \textit{83.66$\pm$0.27} & \textit{13.15$\pm$0.45} & 5.51$\pm$0.22 \\

\hline

\end{tabular}
}
\end{table*}

\begin{table*}[!t]
\centering
\caption{Efficiency and average performance across six datasets. Models were trained from scratch in nnU-Net~\cite{nnunet} using official, publicly available implementations for the compared methods. Latency is reported as mean$\pm$standard deviation over 50 runs with $3\times256\times256$ inputs on an Intel Core i9-13980HX CPU and an NVIDIA GeForce RTX 4080 Laptop GPU. Modality-macro Dice and IoU first average ISIC-2017 and ISIC-2018 into a single dermoscopy score and then assign equal weight to dermoscopy, endoscopy, fundus photography, ultrasound, and microscopy. Mean Dice rank is computed across the six datasets, with average ranks
assigned to tied results.}

\label{tab:ultralight_comparison}

\scriptsize
\setlength{\tabcolsep}{2.5pt}
\renewcommand{\arraystretch}{1.05}

\resizebox{\textwidth}{!}{
\begin{tabular}{@{}llc|cc|cc|c|cc|c@{}}
\hline

Arch.
& Model
& Year
& Params
& GFLOPs$\downarrow$
& \multicolumn{2}{c|}{Latency (ms)$\downarrow$}
& Peak VRAM
& \multicolumn{2}{c|}{Modality-macro}
& Mean Dice
\\

\cline{6-7}
\cline{9-10}

&
&
& (M)$\downarrow$
&
& i9-13980HX
& RTX 4080
& (GB)$\downarrow$
& Dice$\uparrow$
& IoU$\uparrow$
& rank$\downarrow$
\\

\hline

\multirow{16}{*}{Conv}

& U$^{2}$Net \cite{u2net}
& 2020 & 44.024 & 37.696
& 627.72$\pm$7.93
& 7.49$\pm$0.01
& 0.341
& 86.56 & 78.86 & 4.67 \\

& nnU-Net \cite{nnunet}
& 2021 & 33.472 & 14.882
& 231.60$\pm$1.29
& 2.85$\pm$0.01
& 0.193
& 86.58 & 78.74 & 4.58 \\

& UNeXt \cite{unext}
& 2022 & 1.472 & 0.577
& 21.19$\pm$0.23
& \textbf{1.13$\pm$0.04}
& 0.026
& 85.52 & 77.57 & 11.08 \\

& MALUNet \cite{malunet}
& 2022 & 0.178 & 0.091
& 11.06$\pm$0.26
& 2.90$\pm$0.56
& 0.014
& 84.65 & 76.13 & 13.83 \\

& EGE-UNet \cite{egeunet}
& 2023 & 0.053 & \textit{0.079}
& 12.20$\pm$0.26
& 3.91$\pm$0.94
& 0.010
& 84.13 & 75.55 & 18.00 \\

& CMUNet \cite{cmunet}
& 2023 & 49.932 & 91.108
& 1313.04$\pm$12.03
& 10.95$\pm$0.03
& 0.344
& 86.15 & 78.43 & 8.08 \\

& LB-UNet \cite{lbunet}
& 2024 & 0.056 & 0.103
& \textbf{7.92$\pm$0.23}
& 2.39$\pm$0.06
& \textit{0.008}
& 84.51 & 76.04 & 16.50 \\

& CMUNeXt-S \cite{cmunext}
& 2024 & 0.418 & 1.049
& 44.81$\pm$2.71
& \textit{1.25$\pm$0.02}
& 0.028
& 85.75 & 77.87 & 10.42 \\

& TinyU-Net \cite{tinyunet}
& 2024 & 0.481 & 1.614
& 185.71$\pm$1.36
& 2.85$\pm$0.02
& 0.142
& 85.79 & 78.04 & 8.25 \\

& AULUNet \cite{aulunet}
& 2025 & \textit{0.029} & \textbf{0.069}
& 9.23$\pm$0.63
& 1.38$\pm$0.05
& 0.012
& 84.67 & 76.38 & 14.50 \\

& MK-UNet \cite{mkunet}
& 2025 & 0.316 & 0.310
& 31.02$\pm$0.44
& 2.49$\pm$0.09
& 0.030
& 84.73 & 76.31 & 15.83 \\

& UltraSeg130K \cite{ultraseg}
& 2026 & 0.148 & 0.155
& 10.83$\pm$0.19
& 2.17$\pm$0.06
& 0.016
& 84.21 & 75.66 & 14.58 \\

& DHRNet \cite{dhrnet}
& 2026 & 0.065 & 0.081
& \textit{8.33$\pm$0.20}
& 2.40$\pm$0.05
& \textbf{0.005}
& 84.75 & 76.31 & 13.42 \\

& \textbf{LightMIS-T (ours)}
& -- & \textbf{0.017} & 0.099
& 13.99$\pm$0.24
& 2.31$\pm$0.04
& 0.012
& 85.92 & 77.87 & 9.08 \\

& \textbf{LightMIS-S (ours)}
& -- & 0.038 & 0.195
& 20.21$\pm$0.29
& 2.34$\pm$0.05
& 0.018
& 86.30 & 78.45 & 6.75 \\

& \textbf{LightMIS (ours)}
& -- & 0.131 & 0.575
& 41.83$\pm$0.63
& 2.36$\pm$0.04
& 0.037
& \textit{86.71} & \textit{78.99} & \textbf{3.42} \\

\hline

Trans
& SwinUNet \cite{swinunet}
& 2022 & 27.176 & 8.093
& 217.08$\pm$1.41
& 3.49$\pm$0.01
& 0.168
& 82.31 & 73.55 & 18.00 \\

\hline

\multirow{4}{*}{Hybrid}

& TransAttUNet \cite{transattunet}
& 2024 & 25.966 & 88.763
& 1385.54$\pm$7.56
& 9.86$\pm$0.01
& 0.384
& 86.20 & 78.23 & 10.67 \\

& TransUNet \cite{transunet}
& 2024 & 91.719 & 30.286
& 442.60$\pm$2.95
& 5.90$\pm$0.06
& 0.381
& 80.83 & 71.67 & 20.83 \\

& nnWNet \cite{nnwnet}
& 2025 & 7.044 & 10.618
& 214.64$\pm$1.74
& 5.03$\pm$0.01
& 0.116
& 86.51 & 78.84 & 4.50 \\

& Mobile U-ViT \cite{mobileuvit}
& 2025 & 1.390 & 3.294
& 138.53$\pm$6.48
& 2.51$\pm$0.09
& 0.056
& \textbf{86.75} & \textbf{79.07} & \textit{4.00} \\

\hline
\end{tabular}
}
\end{table*}

\begin{table*}[!t]
\caption{Controlled versus recommended configurations on DRIVE and BUSI. Dice and IoU are reported as mean$\pm$standard deviation over five folds. Controlled configurations follow the common nnU-Net-based protocol~\cite{nnunet} used by nnWNet~\cite{nnwnet}. Recommended configurations change only the optimizer, initial learning rate, scheduler, and weight decay according to the corresponding papers.}

\label{tab:controlled_recommended}

\centering
\footnotesize
\setlength{\tabcolsep}{2.5pt}
\renewcommand{\arraystretch}{1.08}

\begin{tabular*}{\textwidth}{
@{\extracolsep{\fill}}
l|cc|cc|cc|cc
@{}
}
\hline

\multirow{3}{*}{Model}
& \multicolumn{4}{c|}{DRIVE}
& \multicolumn{4}{c}{BUSI}
\\

\cline{2-9}

& \multicolumn{2}{c|}{Dice$\uparrow$}
& \multicolumn{2}{c|}{IoU$\uparrow$}
& \multicolumn{2}{c|}{Dice$\uparrow$}
& \multicolumn{2}{c}{IoU$\uparrow$}
\\

\cline{2-9}

& Controlled & Recommended
& Controlled & Recommended
& Controlled & Recommended
& Controlled & Recommended
\\

\hline

UNeXt~\cite{unext} & 81.26$\pm$0.92 & 80.34$\pm$1.09 & 68.55$\pm$1.16 & 67.27$\pm$1.35 & 78.16$\pm$2.33 & 76.00$\pm$1.48 & 70.17$\pm$2.02 & 67.63$\pm$1.10 \\
MALUNet~\cite{malunet} & 78.67$\pm$1.29 & 79.11$\pm$1.09 & 64.98$\pm$1.56 & 65.59$\pm$1.29 & 78.70$\pm$1.87 & 78.59$\pm$0.93 & 70.44$\pm$1.77 & 70.23$\pm$0.96 \\
EGE-UNet~\cite{egeunet} & 79.78$\pm$1.27 & 79.98$\pm$0.71 & 66.54$\pm$1.54 & 66.75$\pm$0.85 & 77.25$\pm$1.22 & 78.04$\pm$1.64 & 68.62$\pm$1.22 & 69.31$\pm$1.42 \\
CMUNet~\cite{cmunet} & 81.41$\pm$1.27 & 81.48$\pm$0.84 & 68.79$\pm$1.61 & 68.85$\pm$1.10 & 78.36$\pm$1.56 & 79.51$\pm$1.68 & 70.64$\pm$1.47 & 71.87$\pm$1.45 \\
LB-UNet~\cite{lbunet} & 79.70$\pm$1.33 & 79.99$\pm$0.97 & 66.44$\pm$1.58 & 66.81$\pm$1.13 & 77.52$\pm$1.43 & 77.12$\pm$2.23 & 69.08$\pm$1.11 & 68.51$\pm$1.98 \\
CMUNeXt-S~\cite{cmunext} & 81.75$\pm$1.03 & 81.85$\pm$1.14 & 69.26$\pm$1.31 & 69.42$\pm$1.45 & 78.58$\pm$1.69 & 78.02$\pm$1.91 & 70.65$\pm$1.54 & 70.19$\pm$1.56 \\
TinyU-Net \cite{tinyunet} & 81.75$\pm$0.96 & 81.52$\pm$0.92 & 69.27$\pm$1.20 & 68.93$\pm$1.15 & 77.59$\pm$2.57 & 75.20$\pm$1.66 & 69.83$\pm$2.41 & 67.03$\pm$1.33 \\
AULUNet~\cite{aulunet} & 80.59$\pm$0.95 & 80.17$\pm$1.43 & 67.65$\pm$1.15 & 67.16$\pm$1.61 & 77.91$\pm$2.20 & 74.26$\pm$2.68 & 69.71$\pm$2.06 & 65.46$\pm$2.37 \\
MK-UNet~\cite{mkunet} & 80.06$\pm$1.45 & 81.38$\pm$1.35 & 66.89$\pm$1.84 & 68.74$\pm$1.76 & 77.56$\pm$0.79 & 78.81$\pm$1.65 & 68.89$\pm$0.87 & 70.79$\pm$1.46 \\
DHRNet~\cite{dhrnet} & 78.27$\pm$1.40 & 78.85$\pm$1.50 & 64.45$\pm$1.68 & 65.25$\pm$1.79 & 78.36$\pm$2.08 & 79.10$\pm$1.63 & 70.17$\pm$1.85 & 70.81$\pm$1.37 \\
UltraSeg130K~\cite{ultraseg} & 78.72$\pm$1.67 & 78.97$\pm$1.24 & 65.13$\pm$1.96 & 65.41$\pm$1.43 & 77.98$\pm$2.06 & 76.27$\pm$1.77 & 69.62$\pm$1.94 & 67.53$\pm$1.54 \\
Mobile U-ViT~\cite{mobileuvit} & 81.63$\pm$0.97 & 81.82$\pm$0.95 & 69.10$\pm$1.20 & 69.37$\pm$1.19 & 80.90$\pm$1.18 & 79.88$\pm$0.85 & 73.27$\pm$0.83 & 71.96$\pm$0.67 \\

\hline
\end{tabular*}

\end{table*}

We compare LightMIS with convolutional, Transformer-based, and hybrid medical image segmentation architectures, including recent lightweight models, across all six datasets. To ensure a controlled and reproducible comparison, we include only methods with official, publicly available implementations. All models are trained from scratch without pretrained weights using the same nnU-Net-based training and evaluation protocol. Dataset-specific Dice, IoU, HD95, and ASSD results are reported in Tables~\ref{tab:results_kvasir_drive}, ~\ref{tab:results_busi_dsb18}, and~\ref{tab:results_isic}, whereas Table~\ref{tab:ultralight_comparison} summarizes modality-macro performance across five imaging modalities represented by the six datasets, model complexity, inference latency, and peak GPU memory usage. Latency was measured from model invocation to output generation for a batch size of 1 with 3 $\times$ 256 $\times$ 256 inputs, using \texttt{time.perf\_counter\_ns()} on CPU and CUDA events on GPU, with model loading and device transfers excluded, whereas peak VRAM was recorded using \texttt{torch.cuda.max\_memory\_allocated()} during GPU inference.  Dice and IoU are first averaged over the five folds for each dataset. The ISIC-2017 and ISIC-2018 dataset means are then averaged into a single dermoscopy score, after which the five modality-level scores are weighted equally to obtain the unweighted modality-macro averages. For a consistent complexity comparison, all parameter counts and GFLOPs were recomputed locally using the same measurement procedure. Trainable parameters were counted directly from the instantiated PyTorch models, whereas GFLOPs were computed for every architecture using \texttt{fvcore.nn.FlopCountAnalysis} with an input tensor of size ($1\times3\times256\times256$), following the protocol adopted by the NTIRE Efficient Super-Resolution challenges~\cite{ren2024ntireesr,ren2025ntireesr,ren2026ntireesr}. Consequently, the recomputed values may differ from those reported in the original publications due to differences in counting conventions.

\begin{figure*}[!t]
\centering

\setlength{\tabcolsep}{0pt}
\renewcommand{\arraystretch}{0}

\newlength{\datasetcol}
\setlength{\datasetcol}{0.08\textwidth}

\newlength{\imagecol}
\setlength{\imagecol}{0.115\textwidth}

\newcommand{\qualrowsep}{
  \noalign{
    \begingroup
    \color{qualblue}
    \hrule height 1pt
    \endgroup
  }
}

\makebox[\textwidth][c]{
\begin{tabular}{
@{}
>{\centering\arraybackslash}m{\datasetcol}
!{\color{qualblue}\vrule width 1pt}
>{\centering\arraybackslash}m{\imagecol}
!{\color{qualblue}\vrule width 1pt}
>{\centering\arraybackslash}m{\imagecol}
!{\color{qualblue}\vrule width 1pt}
>{\centering\arraybackslash}m{\imagecol}
!{\color{qualblue}\vrule width 1pt}
>{\centering\arraybackslash}m{\imagecol}
!{\color{qualblue}\vrule width 1pt}
>{\centering\arraybackslash}m{\imagecol}
!{\color{qualblue}\vrule width 1pt}
>{\centering\arraybackslash}m{\imagecol}
!{\color{qualblue}\vrule width 1pt}
>{\centering\arraybackslash}m{\imagecol}
!{\color{qualblue}\vrule width 1pt}
>{\centering\arraybackslash}m{\imagecol}
@{}
}

\textbf{Dataset}
&
\textbf{Input}
&
\shortstack{\textbf{Ground}\\\textbf{Truth}}
&
\textbf{AULUNet}
&
\shortstack{\textbf{Mobile}\\\textbf{U-ViT}}
&
\textbf{nnWNet}
&
\textbf{nnU-Net}
&
\textbf{U$^{2}$Net}
&
\shortstack{\textbf{LightMIS}\\\textbf{(ours)}}
\\
\qualrowsep

\textbf{DRIVE} case039
&
\includegraphics[draft=false,width=\linewidth]
{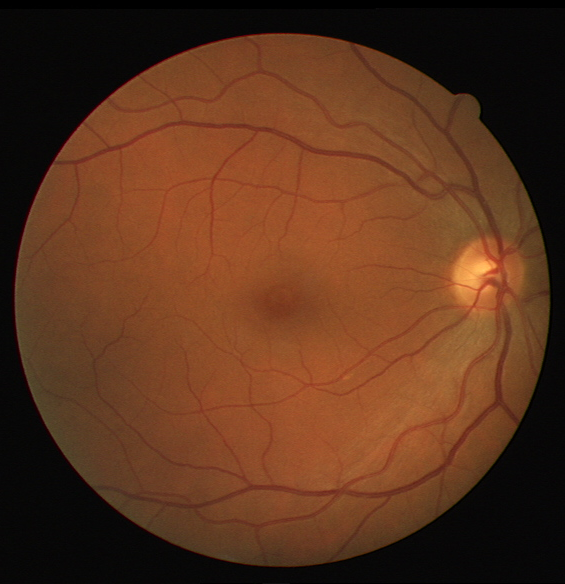}
&
\includegraphics[draft=false,width=\linewidth]
{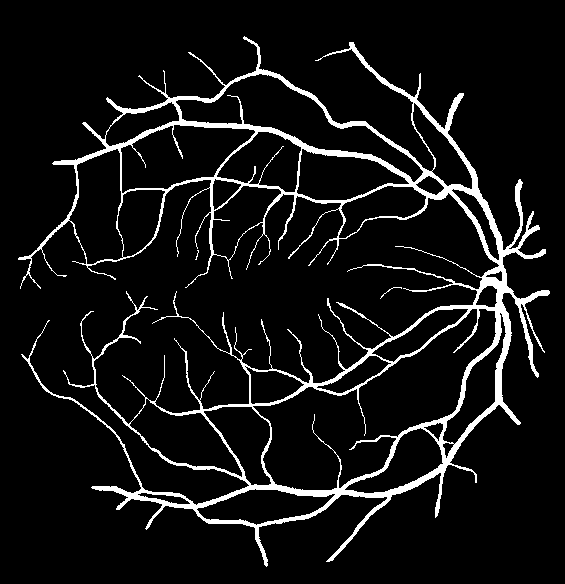}
&
\includegraphics[draft=false,width=\linewidth]
{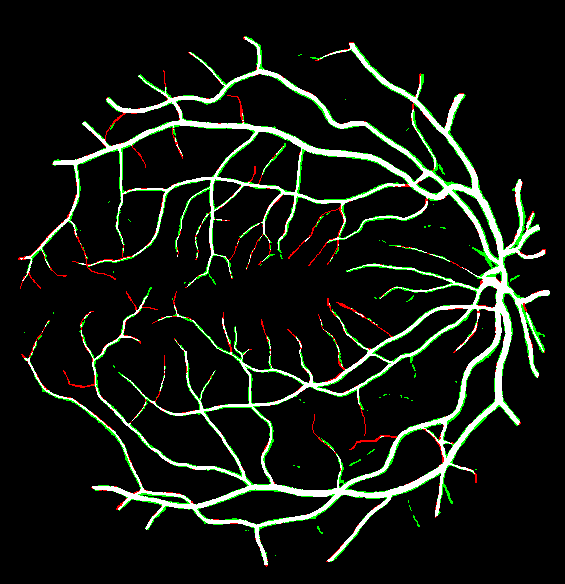}
&
\includegraphics[draft=false,width=\linewidth]
{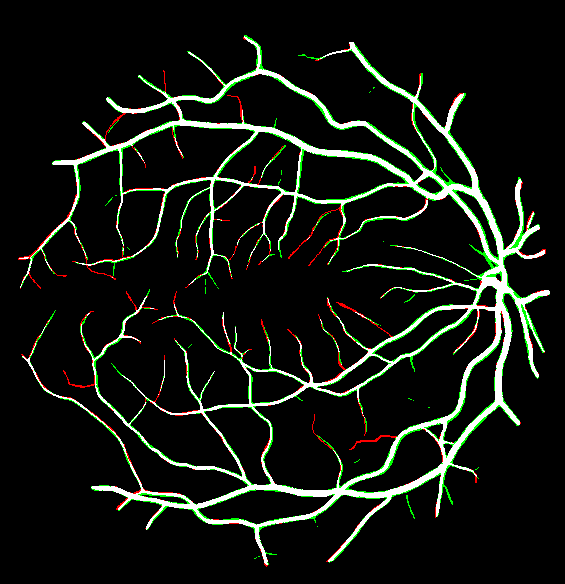}
&
\includegraphics[draft=false,width=\linewidth]
{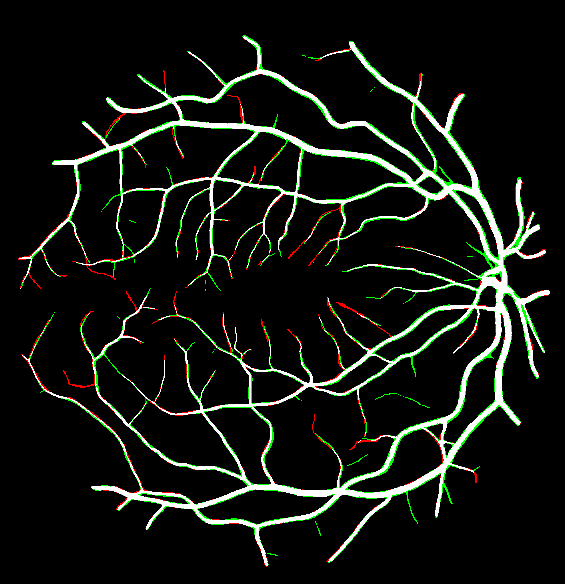}
&
\includegraphics[draft=false,width=\linewidth]
{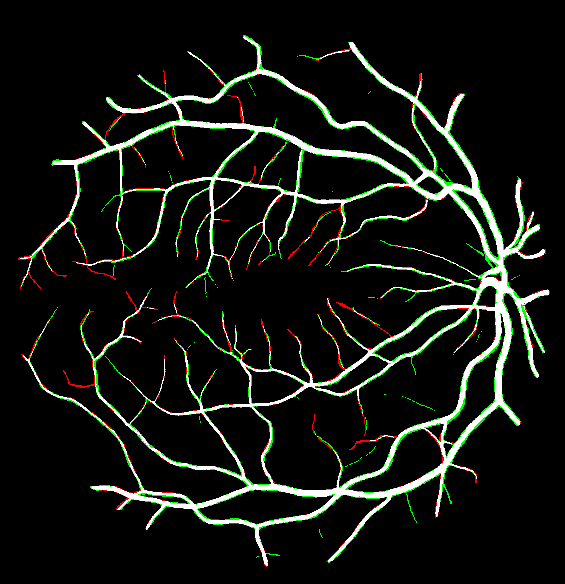}
&
\includegraphics[draft=false,width=\linewidth]
{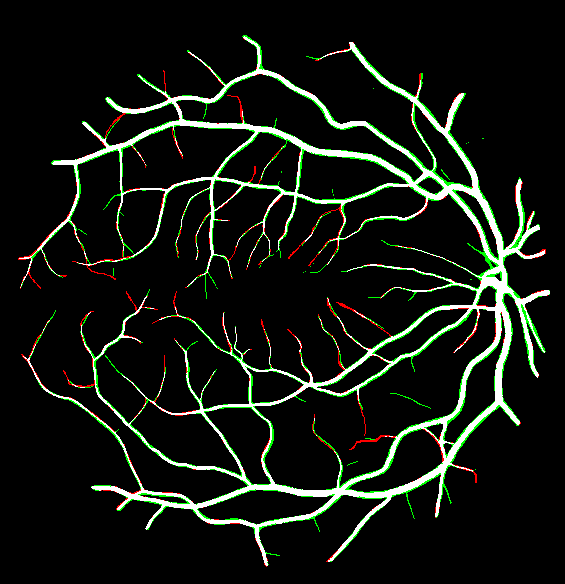}
&
\includegraphics[draft=false,width=\linewidth]
{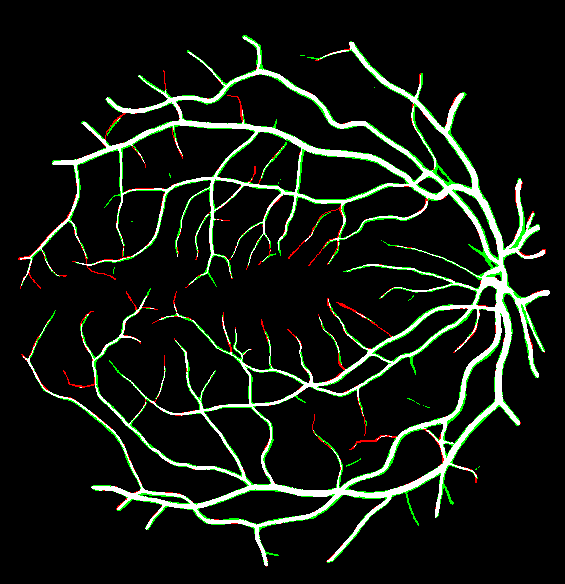}
\\
\qualrowsep

\textbf{Kvasir-SEG} case952
&
\includegraphics[draft=false,width=\linewidth]
{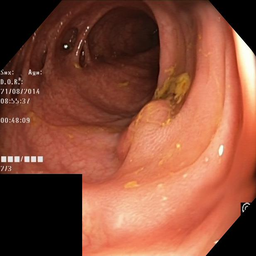}
&
\includegraphics[draft=false,width=\linewidth]
{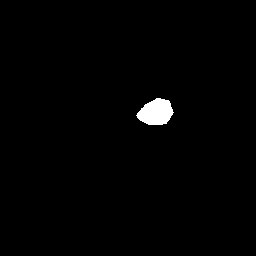}
&
\includegraphics[draft=false,width=\linewidth]
{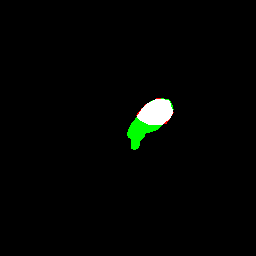}
&
\includegraphics[draft=false,width=\linewidth]
{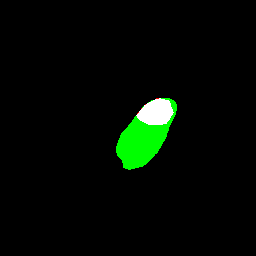}
&
\includegraphics[draft=false,width=\linewidth]
{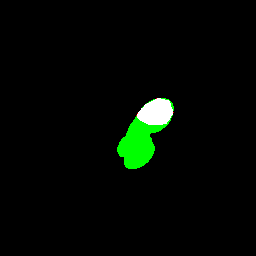}
&
\includegraphics[draft=false,width=\linewidth]
{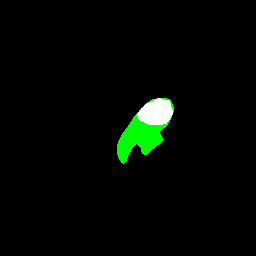}
&
\includegraphics[draft=false,width=\linewidth]
{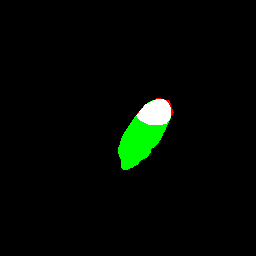}
&
\includegraphics[draft=false,width=\linewidth]
{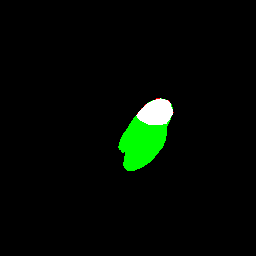}
\\
\qualrowsep

\textbf{DSB18} case654
&
\includegraphics[draft=false,width=\linewidth]
{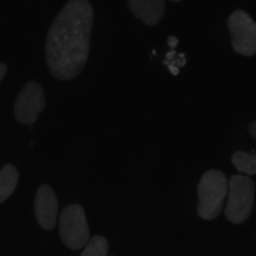}
&
\includegraphics[draft=false,width=\linewidth]
{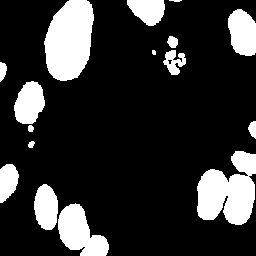}
&
\includegraphics[draft=false,width=\linewidth]
{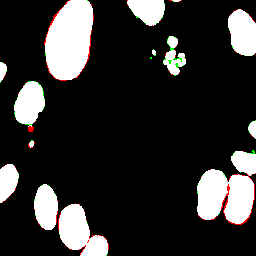}
&
\includegraphics[draft=false,width=\linewidth]
{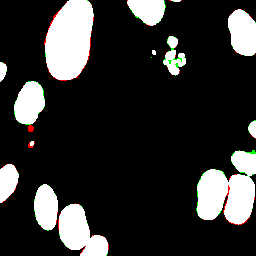}
&
\includegraphics[draft=false,width=\linewidth]
{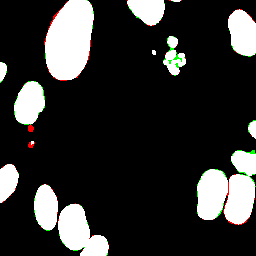}
&
\includegraphics[draft=false,width=\linewidth]
{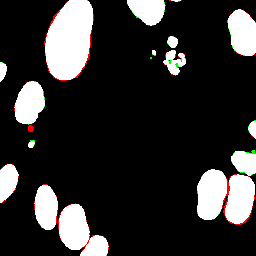}
&
\includegraphics[draft=false,width=\linewidth]
{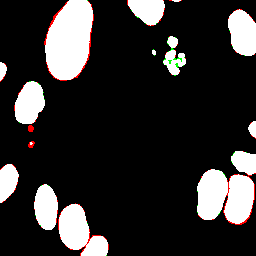}
&
\includegraphics[draft=false,width=\linewidth]
{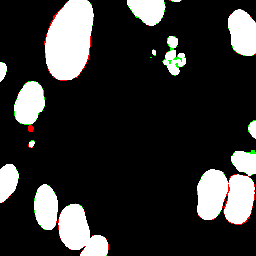}
\\
\qualrowsep

\textbf{BUSI} case612
&
\includegraphics[draft=false,width=\linewidth]
{images/qualitative/BUSI/Dataset250_BUSI_case612_00_input.png}
&
\includegraphics[draft=false,width=\linewidth]
{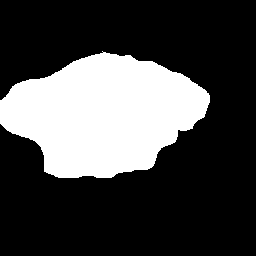}
&
\includegraphics[draft=false,width=\linewidth]
{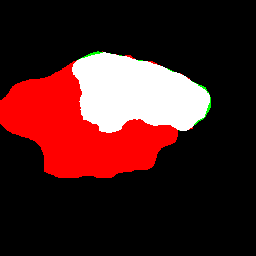}
&
\includegraphics[draft=false,width=\linewidth]
{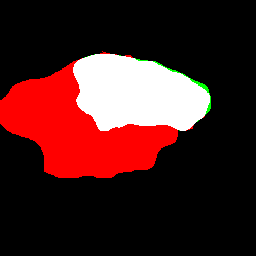}
&
\includegraphics[draft=false,width=\linewidth]
{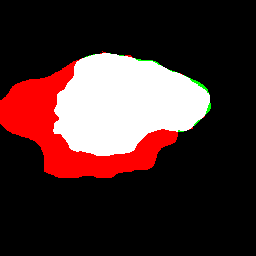}
&
\includegraphics[draft=false,width=\linewidth]
{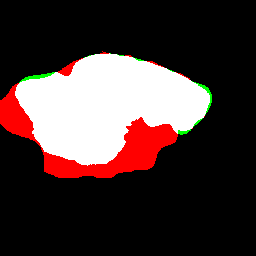}
&
\includegraphics[draft=false,width=\linewidth]
{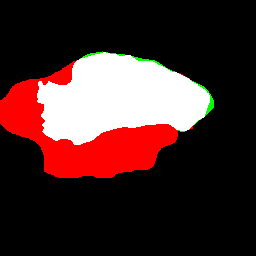}
&
\includegraphics[draft=false,width=\linewidth]
{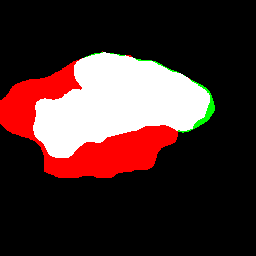}
\\
\qualrowsep

\textbf{ISIC-2017} case992
&
\includegraphics[draft=false,width=\linewidth]
{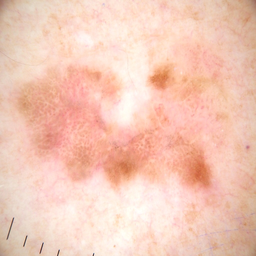}
&
\includegraphics[draft=false,width=\linewidth]
{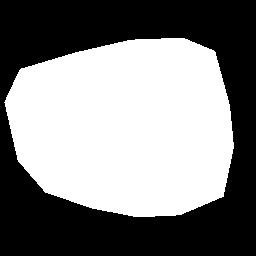}
&
\includegraphics[draft=false,width=\linewidth]
{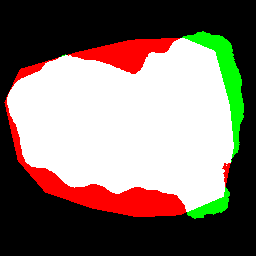}
&
\includegraphics[draft=false,width=\linewidth]
{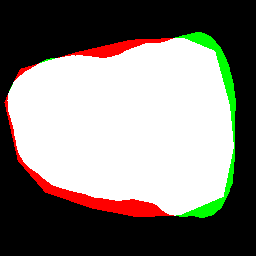}
&
\includegraphics[draft=false,width=\linewidth]
{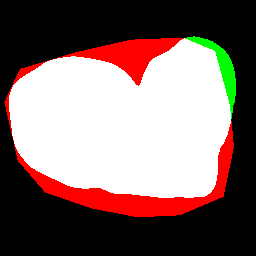}
&
\includegraphics[draft=false,width=\linewidth]
{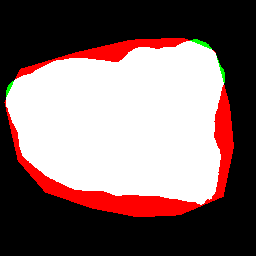}
&
\includegraphics[draft=false,width=\linewidth]
{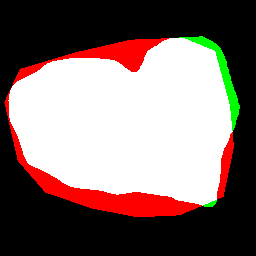}
&
\includegraphics[draft=false,width=\linewidth]
{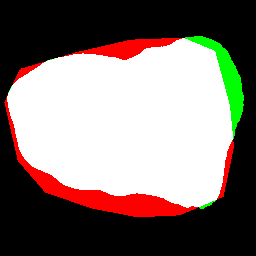}
\\
\qualrowsep

\textbf{ISIC-2018} case998
& 
\includegraphics[draft=false,width=\linewidth]
{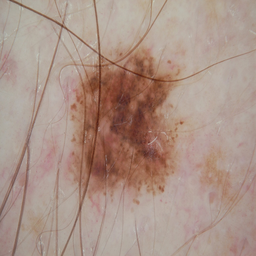}
&
\includegraphics[draft=false,width=\linewidth]
{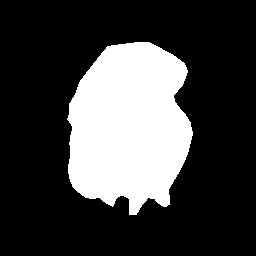}
&
\includegraphics[draft=false,width=\linewidth]
{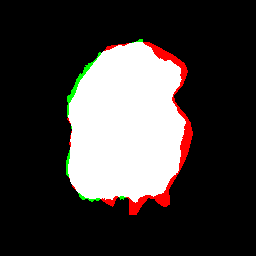}
&
\includegraphics[draft=false,width=\linewidth]
{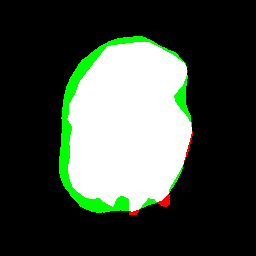}
&
\includegraphics[draft=false,width=\linewidth]
{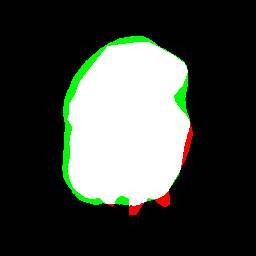}
&
\includegraphics[draft=false,width=\linewidth]
{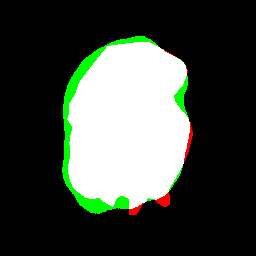}
&
\includegraphics[draft=false,width=\linewidth]
{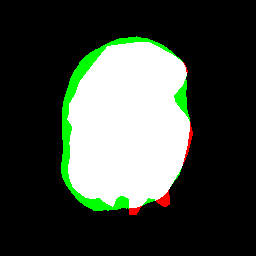}
&
\includegraphics[draft=false,width=\linewidth]
{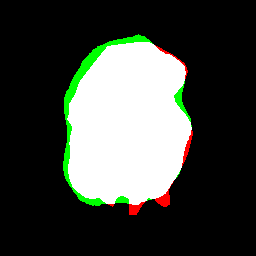}
\\

\end{tabular}
}

\caption{Qualitative comparison across six medical image segmentation datasets. White, red, green, and black denote true positives, false negatives, false positives, and true negatives, respectively. The case identifier for each image is shown below the corresponding dataset. Identifiers were assigned according to the lexicographical order of the original filenames. Images were selected prospectively.}
\label{fig:qualitative_comparison}

\end{figure*}

LightMIS demonstrates strong and competitive performance across the evaluated datasets. On DRIVE, it achieves the highest Dice and IoU scores of 82.15\% and 69.82\%, respectively. This result indicates that LightMIS effectively preserves the fine multi-scale structures required for delineating thin and highly branched retinal vessels. On BUSI, LightMIS achieves 80.16\% Dice and 72.28\% IoU, which are 0.44 and 0.19 percentage points higher than those of nnWNet, respectively, whereas Mobile U-ViT achieves the best results on this dataset, with 80.90\% Dice and 73.27\% IoU. LightMIS reaches 89.32\% Dice and 82.60\% IoU on ISIC-2017 and 89.83\% Dice and 83.39\% IoU on ISIC-2018. On DSB18, it obtains 91.81\% Dice and 85.62\% IoU, which are 0.29 and 0.40 percentage points below the respective highest scores, while achieving the lowest HD95 and ASSD values of 4.59 and 1.53. On Kvasir-SEG, LightMIS performs competitively with substantially larger architectures. nnU-Net achieves a Dice score of 90.02\%, only 0.15 percentage points above the score obtained by LightMIS. On the same dataset, LightMIS also exceeds nnWNet, which achieves 89.69\% Dice, and obtains the lowest ASSD of 5.44. Furthermore, LightMIS outperforms several recent lightweight models, including MK-UNet, AULUNet, LB-UNet, MALUNet, and EGE-UNet, whose Dice scores range from 84.08\% to 85.75\% on Kvasir-SEG dataset.

LightMIS achieves the second-highest observed modality-macro Dice and IoU scores and the best mean Dice rank across the six datasets in Table~\ref{tab:ultralight_comparison}. Mobile U-ViT obtains the highest macro-averaged Dice and IoU scores of 86.75\% and 79.07\%, respectively, closely followed by LightMIS with 86.71\% Dice and 78.99\% IoU. LightMIS also exceeds nnWNet by 0.20 and 0.15 percentage points in Dice and IoU, respectively, and exceeds nnU-Net by 0.13 and 0.25 percentage points for the same metrics. We therefore interpret these values as close observed averages rather than evidence of statistical superiority. Paired confidence intervals are reported in Section~\ref{sec:statistics}. Despite the comparable observed performance, LightMIS uses 90.58\% fewer parameters and 82.54\% fewer GFLOPs than Mobile U-ViT, as well as 98.14\% fewer parameters and 94.58\% fewer GFLOPs than nnWNet. Compared with the nnU-Net baseline, LightMIS uses 99.61\% fewer parameters and 96.14\% fewer GFLOPs. LightMIS also records lower inference latency and peak GPU memory usage than all three models, demonstrating a favorable balance between segmentation performance and computational requirements. Compared with TinyU-Net, it uses 72.77\% fewer parameters and 64.37\% fewer GFLOPs while improving the macro-average Dice and IoU by 0.92 and 0.95 percentage points, respectively. Relative to CMUNeXt-S, LightMIS reduces the parameter count by 68.66\% and GFLOPs by 45.19\%, with corresponding improvements of 0.96 and 1.12 percentage points.

\begin{figure}[!t]
    \centering
    \includegraphics[width=\columnwidth]
    {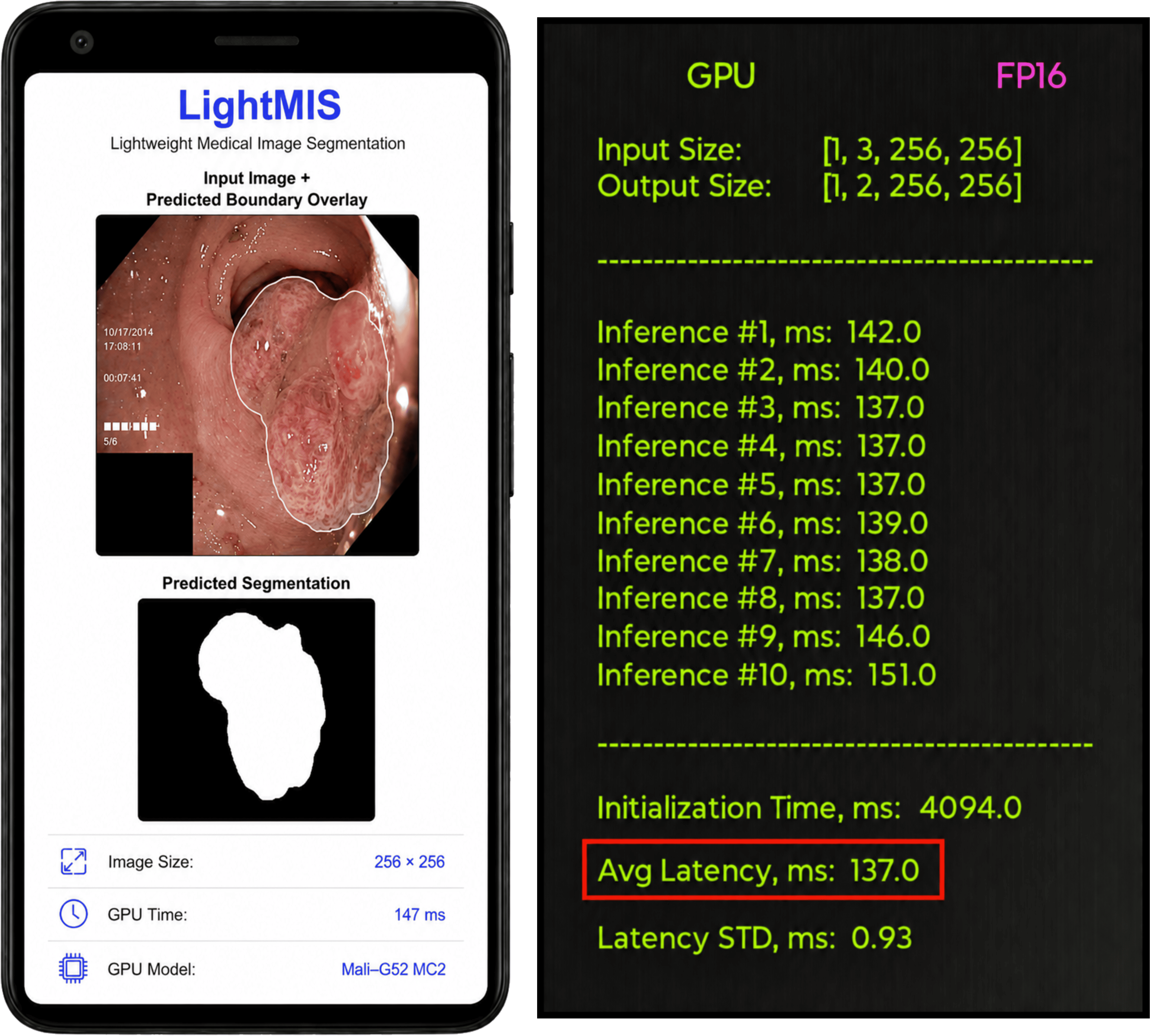}
    \caption{(L) Sample medical image segmentation application on a mobile edge device using LightMIS. (R) LightMIS GPU latency measured locally on the smartphone using AI Benchmark \cite{aibenchmark}.}
    \label{fig:local_benchmark}
\end{figure}

The observed performance of some architectures varies considerably across imaging domains. MALUNet achieves Dice scores of 89.11\% and 89.48\% on ISIC-2017 and ISIC-2018, respectively, but obtains 78.67\% on DRIVE. Similarly, MK-UNet achieves 91.25\% Dice on DSB18 but 85.72\% on Kvasir-SEG, whereas EGE-UNet and AULUNet obtain 84.08\% and 84.60\%, respectively, on Kvasir-SEG. These findings suggest that strong performance within one imaging domain does not necessarily translate into equally strong performance across other segmentation tasks.

Under the common from-scratch training protocol, SwinUNet and TransUNet exhibit substantially lower performance on some datasets. SwinUNet obtains Dice scores of 80.15\% on Kvasir-SEG and 71.45\% on BUSI, whereas TransUNet achieves 77.69\% and 71.07\%, respectively. This pattern is consistent with the unified benchmark reported by nnWNet~\cite{nnwnet}, in which SwinUNet performed below several convolutional and hybrid models on DRIVE, whereas no valid result was reported for two datasets. Nevertheless, some values reported by nnWNet for the same models and datasets show variations relative to our results. These variations may be attributable to several implementation-related factors. The dataset-conversion script used in the nnWNet study was not included in the publicly available release. Consequently, it remains unclear whether identical data splits were used in our study. Additionally, nnWNet uses deep supervision, whereas we disabled it for all models in our study, and the PyTorch version is not reported, which may lead to potential software-version effects on model performance.

\begin{table*}[!t]
\caption{On-device delegated inference at batch size one. Mean$\pm$standard deviation, P50, and P95 characterize the interval from CPU-to-GPU transfer at delegate invocation to the corresponding
GPU-to-CPU return. Peak process RSS was sampled at 5\,ms intervals. All measurements used a batch size of one and a fixed $3\times256\times256$ input, with model weights retained in FP32 and delegated operations executed in FP16 through the LiteRT GPU delegate.}
\label{tab:mobile_inference}

\centering
\footnotesize
\setlength{\tabcolsep}{1.2pt}
\renewcommand{\arraystretch}{1.10}

\begin{tabularx}{\linewidth}{
@{}
>{\centering\arraybackslash}p{0.050\linewidth}
>{\raggedright\arraybackslash}X
|
>{\centering\arraybackslash}p{0.060\linewidth}
>{\centering\arraybackslash}p{0.075\linewidth}
|
>{\centering\arraybackslash}p{0.120\linewidth}
>{\centering\arraybackslash}p{0.068\linewidth}
>{\centering\arraybackslash}p{0.068\linewidth}
>{\centering\arraybackslash}p{0.095\linewidth}
>{\centering\arraybackslash}p{0.120\linewidth}
>{\centering\arraybackslash}p{0.085\linewidth}
@{}
}
\hline

\multirow{2}{*}{Arch.}
& \multirow{2}{*}{Model}
& \multirow{2}{*}{Precision}
& \multirow{2}{*}{\shortstack{File size\\(MB)$\downarrow$}}
& \multicolumn{3}{c}{Delegate latency (ms)$\downarrow$}
& \multirow{2}{*}{\shortstack{Peak RSS\\(MB)$\downarrow$}}
& \multirow{2}{*}{\shortstack{GPU coverage\\(\%)}}
& \multirow{2}{*}{\shortstack{CPU fallback\\nodes}}
\\

\cline{5-7}

& & & &
Mean$\pm$SD
& P50$\downarrow$
& P95$\downarrow$
& & &
\\

\hline

\multirow{16}{*}{Conv}

& U$^{2}$Net \cite{u2net}
& FP16 & 176.332
& 703.24$\pm$16.96
& 713.64 & 718.26
& 677.57
& 100\%
& 0
\\

& nnU-Net \cite{nnunet}
& FP16 & 134.263
& 458.77$\pm$13.54
& 450.93 & 481.94
& 487.34
& 100\%
& 0
\\

& UNeXt \cite{unext}
& FP16 & 6.057
& \textit{42.14$\pm$0.54}
& \textit{42.08}
& \textit{43.03}
& 209.05
& 100\%
& 0
\\

& MALUNet \cite{malunet}
& FP16 & 1.495
& 61.19$\pm$3.28
& 60.96 & 67.48
& 214.10
& 94.56\%
& 81
\\

& EGE-UNet \cite{egeunet}
& FP16 & 1.039
& 64.64$\pm$4.29
& 64.83 & 70.24
& 218.55
& 97.49\%
& 21
\\

& CMUNet \cite{cmunet}
& FP16 & 199.935
& 1342.16$\pm$11.46
& 1338.81 & 1363.50
& 898.67
& 100\%
& 0
\\

& LB-UNet \cite{lbunet}
& FP16 & 0.672
& 90.13$\pm$3.34
& 90.60 & 94.59
& 222.06
& 95.11\%
& 38
\\

& CMUNeXt-S \cite{cmunext}
& FP16 & 1.844
& 92.70$\pm$0.57
& 92.70 & 93.61
& 192.15
& 100\%
& 0
\\

& TinyU-Net \cite{tinyunet}
& FP16 & 2.263
& 2635.67$\pm$161.73
& 2547.58 & 2957.42
& 1679.34
& 54.08\%
& 214
\\

& AULUNet \cite{aulunet}
& FP16 & \textbf{0.260}
& \textbf{18.57$\pm$0.40}
& \textbf{18.53}
& \textbf{19.38}
& \textbf{184.60}
& 100\%
& 0
\\

& MK-UNet \cite{mkunet}
& FP16 & 1.547
& 148.35$\pm$3.14
& 148.13 & 152.19
& 219.73
& 99.57\%
& 1
\\

& UltraSeg130K \cite{ultraseg}
& FP16 & 0.947
& 92.84$\pm$3.97
& 93.36 & 98.13
& 221.88
& 97.62\%
& 15
\\

& DHRNet \cite{dhrnet}
& FP16 & 0.805
& 42.45$\pm$2.23
& 42.52 & 46.34
& \textit{188.35}
& 92.35\%
& 83
\\

& \textbf{LightMIS-T (ours)}
& FP16 & \textit{0.579}
& 53.39$\pm$0.58
& 53.31 & 54.65
& 191.98
& 100\%
& 0
\\

& \textbf{LightMIS-S (ours)}
& FP16 & 0.644
& 76.32$\pm$0.58
& 76.22 & 77.42
& 195.68
& 100\%
& 0
\\

& \textbf{LightMIS (ours)}
& FP16 & 1.012
& 138.43$\pm$1.04
& 138.31 & 140.21
& 201.29
& 100\%
& 0
\\

\hline

Trans
& SwinUNet \cite{swinunet}
& FP16 & 112.705
& 1529.27$\pm$16.70
& 1532.19 & 1538.87
& 422.75
& 98.79\%
& 13
\\

\hline

\multirow{3}{*}{Hybrid}

& TransUNet \cite{transunet}
& FP16 & 367.231
& 4717.93$\pm$39.74
& 4710.39 & 4783.19
& 910.66
& 99.84\%
& 1
\\

& nnWNet \cite{nnwnet}
& FP16 & 28.819
& 1453.26$\pm$15.56
& 1454.44 & 1472.07
& 344.87
& 95.99\%
& 45
\\

& Mobile U-ViT \cite{mobileuvit}
& FP16 & 6.027
& 654.58$\pm$23.45
& 657.20 & 686.03
& 233.05
& 99.23\%
& 4
\\

\hline
\end{tabularx}

\end{table*}

\begin{table*}[!t]
\centering
\caption{Operator substitutions used during LiteRT conversion and their numerical validation. Maximum error is computed over all output elements, validation cases, folds, and datasets. Dice change represents the largest absolute dataset-level change across the six datasets.}
\label{tab:conversion}

\footnotesize
\setlength{\tabcolsep}{1.2pt}
\renewcommand{\arraystretch}{1.10}

\begingroup

\renewcommand{\tabularxcolumn}[1]{m{#1}}

\newcommand{\conversionheader}[1]{
    \parbox[c][2.7\baselineskip][c]{\linewidth}{
        \centering #1\par
    }
}

\newcommand{\conversionmodels}{%
    \parbox[c][2.7\baselineskip][c]{\linewidth}{
        \raggedright
        DHRNet~\cite{dhrnet}\\[1pt]
        MK-UNet~\cite{mkunet}\par
    }
}

\begin{tabularx}{\linewidth}{
@{}
>{\raggedright\arraybackslash}m{0.15\linewidth}
|
>{\hsize=0.95\hsize\linewidth=\hsize
  \raggedright\arraybackslash}X
>{\hsize=1.15\hsize\linewidth=\hsize
  \raggedright\arraybackslash}X
>{\hsize=0.90\hsize\linewidth=\hsize
  \raggedright\arraybackslash}X
|
>{\centering\arraybackslash}m{0.10\linewidth}
>{\centering\arraybackslash}m{0.10\linewidth}
@{}
}
\hline

\conversionheader{Model}
&
\conversionheader{Original operator}
&
\conversionheader{LiteRT formulation}
&
\conversionheader{Reason}
&
\conversionheader{Maximum\\error}
&
\conversionheader{Dice change\\(pp)}
\\

\hline

\conversionmodels
&
Adaptive global max pooling
&
Fixed-size max pooling
&
Unsupported dynamic form
&
0
&
0
\\

\hline

MALUNet~\cite{malunet}
&
Explicit mask expansion
&
Broadcast multiplication
&
Avoid materialized expansion
&
0
&
0
\\[2pt]

\hline

Mobile U-ViT~\cite{mobileuvit}
&
Depthwise transposed convolution
&
Grouped $1{\times}1$ convolution followed by reshape and permutation
&
Delegate compatibility
&
$1.17{\times}10^{-2}$
&
$4.94{\times}10^{-4}$
\\[2pt]

\hline
\end{tabularx}

\endgroup
\end{table*}

Although some architectures may achieve stronger results under their original model-specific configurations, their relative performance can change under a unified protocol. Table~\ref{tab:controlled_recommended} compares the Dice and IoU scores achieved by several evaluated models under the adopted controlled nnU-Net-based protocol with those obtained by changing, for each model, only four trainer-level settings according to the original publication: the optimizer, initial learning rate, scheduler, and weight decay. The largest improvements are observed for MK-UNet, whose recommended configuration increases Dice by 1.32 points and IoU by 1.85 points on DRIVE, and Dice by 1.25 points and IoU by 1.90 points on BUSI. In contrast, AULUNet exhibits the largest degradation on BUSI, decreasing by 3.65 Dice points and 4.25 IoU points. Notable degradations on BUSI are also observed for TinyU-Net, which decreases by 2.39 Dice points and 2.80 IoU points. On the same dataset, UNeXt decreases by 2.16 Dice points and 2.54 IoU points, whereas UltraSeg130K decreases by 1.71 Dice points and 2.09 IoU points. This variation across models and datasets suggests that model-specific training configurations do not offer a consistent advantage over the controlled protocol. However, transformer-based models may depend more heavily on pretraining, and training for 200 epochs may affect distinct model families differently. Therefore, the adopted unified protocol provides a controlled comparison but is not necessarily fair to all architectures.

The qualitative comparison in Fig.~\ref{fig:qualitative_comparison} presents one case from each of the six datasets for LightMIS, AULUNet, Mobile U-ViT, nnWNet, nnU-Net, and U$^{2}$Net. AULUNet is included as the closest architectural reference through its AKF block, whereas the remaining methods act as larger high-performing reference architectures. For each dataset, the last case appearing in the split file automatically generated by the nnU-Net framework was selected, corresponding to the final entry in the validation list of the last cross-validation fold. This deterministic rule was applied without inspecting the image content, model predictions, or per-case performance metrics, thereby avoiding performance-based cherry-picking. The corresponding case identifiers, assigned based on the lexicographic ordering of the original filenames, are provided in the figure.

\subsection{On-Device Benchmarking}

Models were converted to LiteRT v1.4.1 and executed on a Motorola Moto G24 Power containing the MediaTek MT6769 SoC, an Arm Mali-G52 MC2 GPU, and 8 GB of memory under Android 14. Fig.~\ref{fig:local_benchmark} provides an illustrative visualization and an example evaluation of LightMIS on the AI Benchmark platform \cite{aibenchmark}, whereas the reported values were obtained in a different environment.  All measurements used a batch size of one and a fixed $3\times256\times256$ input, with model weights retained in FP32 and delegated operations executed in FP16 through the GPU delegate.

The quantitative results reported in Table~\ref{tab:mobile_inference} were obtained using our controlled benchmarking protocol. After 20 warm-up runs, latency was measured over 50 runs in each of 5 independent sessions, with a 100 ms pause between consecutive inferences. Initialization time was excluded. We report median (P50) and 95th-percentile (P95) delegate latency, measured from the CPU-to-GPU transfer at delegate invocation to the corresponding GPU-to-CPU return, delegate coverage, model-file size and maximum process resident set size (RSS), with RSS sampled every $5$\,ms. GPU-delegate coverage was verified for every model, and any CPU fallback is reported.

The LiteRT interpreter used four CPU threads and FP16 inference, with the device charging and its battery level at 100\%, whereas the battery temperature remained below $33\,^{\circ}\mathrm{C}$. For operations not directly supported by LiteRT, equivalent formulations were used as documented in Table~\ref{tab:conversion}. For Mobile U-ViT, differences in floating-point accumulation order introduce a nonzero numerical error, although the maximum absolute change in five-fold mean Dice across the six datasets is only $4.94{\times}10^{-4}$ percentage points. AULUNet~\cite{aulunet} achieves the lowest measured on-device mean latency of $18.57\pm0.40$~ms. This efficiency, together with the adaptive feature-processing capability of its AKF block, motivated the adapted use of AKF within LightMIS. LightMIS-T, LightMIS-S, and LightMIS require $53.39\pm0.58$, $76.32\pm0.58$, and $138.43\pm1.04$~ms, respectively. The full model remains substantially faster than Mobile U-ViT at $654.58\pm23.45$~ms and nnWNet at $1453.26\pm15.56$~ms on the same device. TransAttUNet has no reported mobile latency because its LiteRT model could not be executed by the GPU delegate due to the unsupported \texttt{BROADCAST\_TO} operation.

UNeXt and full LightMIS have nearly identical theoretical computational costs of 0.577 and 0.575 GFLOPs, respectively, and both achieve 100\% GPU coverage. Yet LightMIS is nearly twice as slow on the reported CPU in Table~\ref{tab:ultralight_comparison} and more than three times as slow on the Mali-G52 at 138.43 versus 42.14 ms. Similarly, although neither AULUNet nor LightMIS-T requires any CPU fallback and both are extremely small, AULUNet is much faster on the smartphone, with a P50 latency of 18.53 versus 53.31 ms. In contrast, TinyU-Net achieves only 54.08\% GPU coverage, requires CPU fallback for 214 nodes, and records P50 and P95 latencies of 2547.58 and 2957.42 ms. These discrepancies indicate that GFLOPs alone do not determine real-device efficiency and that branch count, memory movement, kernel-launch overhead, and operator support within the delegate are key determinants of practical latency. These findings position the proposed method as a valuable solution offering a favorable accuracy–efficiency trade-off, rather than as the most hardware-efficient model across all platforms.

\subsection{Statistical Analysis}
\label{sec:statistics}
All model comparisons use paired out-of-fold predictions. For baseline \(m\), dataset \(d\), and evaluation case \(j\), the paired Dice difference is \begin{equation}
\Delta_{m,d,j} = \operatorname{Dice}_{\mathrm{LightMIS},d,j} - \operatorname{Dice}_{m,d,j}.
\end{equation}

Ninety-five percent confidence intervals are estimated using 10,000 paired percentile-bootstrap resamples. Resampling is performed at the image level, separately within each fold, and the five fold-level differences are averaged equally. Two-sided paired sign-flip permutation tests are performed using 100,000 Monte Carlo resamples. The resulting \(p\)-values are adjusted across the 18 comparisons using the Benjamini--Hochberg procedure.

Superiority is confirmed only when the entire confidence interval lies above zero and the adjusted value satisfies \(q_{\mathrm{BH}}<0.05\). No non-inferiority or equivalence analysis is performed, and nonsignificant differences are therefore not interpreted as evidence of comparability. Modality-macro averages are reported separately as descriptive summaries. As shown in Table~\ref{tab:paired_statistics}, among the comparisons with nnU-Net, nnWNet, and Mobile U-ViT, LightMIS achieves statistically confirmed superiority only on DRIVE, where both criteria are satisfied against all three baselines.

\begin{table}[!t]
\centering
\caption{Paired out-of-fold Dice comparisons. Differences are LightMIS minus baseline. Bold requires both a positive 95\% confidence interval and $q_{\mathrm{BH}}<0.05$.}
\label{tab:paired_statistics}

\footnotesize
\setlength{\tabcolsep}{1.2pt}
\renewcommand{\arraystretch}{1.05}

\begin{tabularx}{\columnwidth}{
@{}
>{\raggedright\arraybackslash}p{0.17\columnwidth}
>{\raggedright\arraybackslash}X
>{\centering\arraybackslash}p{0.20\columnwidth}
>{\centering\arraybackslash}p{0.25\columnwidth}
>{\centering\arraybackslash}p{0.14\columnwidth}
@{}
}
\hline

Dataset
& Baseline & $\Delta$ Dice (pp) & 95\% CI & $q_{\mathrm{BH}}$
\\

\hline

\multirow{3}{*}{ISIC-2017}
& Mobile U-ViT
& $-0.269$
& $[-0.550,\ 0.015]$
& 0.2240
\\

& nnU-Net
& $-0.104$
& $[-0.397,\ 0.177]$
& 0.7334
\\

& nnWNet
& $-0.063$
& $[-0.357,\ 0.225]$
& 0.8554
\\

\hline

\multirow{3}{*}{ISIC-2018}
& Mobile U-ViT
& $-0.164$
& $[-0.363,\ 0.035]$
& 0.2639
\\

& nnU-Net
& $-0.193$
& $[-0.432,\ 0.047]$
& 0.2639
\\

& nnWNet
& $-0.014$
& $[-0.221,\ 0.197]$
& 0.9459
\\

\hline

\multirow{3}{*}{Kvasir-SEG}
& Mobile U-ViT
& $+0.221$
& $[-0.239,\ 0.701]$
& 0.7241
\\

& nnU-Net
& $-0.144$
& $[-0.703,\ 0.412]$
& 0.8554
\\

& nnWNet
& $+0.183$
& $[-0.314,\ 0.710]$
& 0.7334
\\

\hline

\multirow{3}{*}{DRIVE}
& Mobile U-ViT
& $\mathbf{+0.518}$
& $\mathbf{[0.311,\ 0.752]}$
& $\mathbf{0.00027}$
\\

& nnU-Net
& $\mathbf{+0.751}$
& $\mathbf{[0.461,\ 1.059]}$
& $\mathbf{0.00018}$
\\

& nnWNet
& $\mathbf{+0.462}$
& $\mathbf{[0.251,\ 0.665]}$
& $\mathbf{0.00150}$
\\

\hline

\multirow{3}{*}{BUSI}
& Mobile U-ViT
& $-0.746$
& $[-1.690,\ 0.161]$
& 0.2639
\\

& nnU-Net
& $-0.005$
& $[-0.952,\ 0.949]$
& 0.9922
\\

& nnWNet
& $+0.435$
& $[-0.629,\ 1.500]$
& 0.7334
\\

\hline

\multirow{3}{*}{DSB18}
& Mobile U-ViT
& $+0.033$
& $[-0.165,\ 0.216]$
& 0.8574
\\

& nnU-Net
& $+0.229$
& $[0.047,\ 0.421]$
& 0.0785
\\

& nnWNet
& $-0.025$
& $[-0.187,\ 0.138]$
& 0.8574
\\

\hline
\end{tabularx}
\end{table}

\begin{table*}[!t]
\caption{LightMIS-T architecture ablation on BUSI and Kvasir-SEG in nnU-Net~\cite{nnunet}. Encoder depth ranges from 2 to 6 levels, with channels $(3,6,10,16,24,40)$ for the 6-level variant, whereas PRF variants use the 5-level LightMIS-T. Results are mean$\pm$SD over five folds.}

\label{tab:arch_ablation}

\centering
\footnotesize
\setlength{\tabcolsep}{2.5pt}
\renewcommand{\arraystretch}{1.08}

\resizebox{\textwidth}{!}{%
\begin{tabular}{@{}l|c|cccc|cccc@{}}
\hline

\multirow{2}{*}{Variant}
& Params 
& \multicolumn{4}{c|}{BUSI}
& \multicolumn{4}{c}{Kvasir-SEG} \\

& (M)
& Dice$\uparrow$
& IoU$\uparrow$
& HD95$\downarrow$
& ASSD$\downarrow$
& Dice$\uparrow$
& IoU$\uparrow$
& HD95$\downarrow$
& ASSD$\downarrow$ \\

\hline

2-level encoder
& 0.004
& 69.97$\pm$1.70 & 59.24$\pm$1.81 & 39.17$\pm$2.35 & 14.18$\pm$1.12
& 71.94$\pm$1.96 & 61.27$\pm$1.83 & 47.10$\pm$4.02 & 15.47$\pm$1.56 \\

3-level encoder
& 0.006
& 75.78$\pm$1.62 & 66.45$\pm$1.51 & 29.38$\pm$1.38 & 10.94$\pm$0.69
& 83.01$\pm$1.39 & 74.87$\pm$1.63 & 30.96$\pm$3.32 & 9.56$\pm$1.28 \\

4-level encoder
& 0.010
& 77.87$\pm$1.56 & 69.57$\pm$1.32 & 24.39$\pm$2.10 & 10.08$\pm$1.58
& 86.76$\pm$1.36 & 79.86$\pm$1.42 & 23.71$\pm$1.73 & 7.04$\pm$0.79 \\

5-level encoder (\textbf{LightMIS-T})
& 0.017
& \textbf{79.62$\pm$1.28} & \textbf{71.48$\pm$0.96} & 23.24$\pm$1.33 & \textbf{8.70$\pm$0.77}
& \textbf{87.52$\pm$1.30} & \textbf{80.96$\pm$1.45} & 22.66$\pm$2.61 & 6.78$\pm$0.99 \\

6-level encoder
& 0.036
& 78.98$\pm$1.57 & 71.02$\pm$1.22 &  23.32$\pm$2.18 & 9.08$\pm$0.91
& 87.40$\pm$1.36 & 80.84$\pm$1.55 &  22.53$\pm$1.77 & 6.59$\pm$0.63 \\

\hline

PRF uniform 3$\times$3 DWConv
& 0.016
& 78.90$\pm$1.85 & 70.81$\pm$1.45 & 23.58$\pm$3.02 & 9.37$\pm$1.63
& 87.19$\pm$2.09 & 80.75$\pm$2.47 & 22.99$\pm$3.68 & 7.01$\pm$1.47 \\

PRF w/o progressive transfer
& 0.017
& 78.27$\pm$1.63 & 70.25$\pm$1.22 & 24.34$\pm$1.94 & 9.84$\pm$1.42
& 87.41$\pm$1.20 & 80.94$\pm$1.47 & \textbf{21.78$\pm$1.96} & \textbf{6.47$\pm$0.79} \\

PRF w/o channel expansion
& 0.014
& 79.60$\pm$1.60 & 71.45$\pm$1.43 & \textbf{22.94$\pm$3.80} & 9.05$\pm$2.34
& 87.11$\pm$1.37 & 80.47$\pm$1.59 & 22.68$\pm$2.82 & 6.81$\pm$1.08 \\

\hline
\end{tabular}
}

\end{table*}

\begin{table*}[!t]
\caption{Scaling analysis of LightMIS across four medical datasets. Configurations I and II decouple encoder channel width from the aggregation-head fusion dimension. Results are mean$\pm$standard deviation over five folds under nnU-Net~\cite{nnunet}.}

\label{tab:scaling_analysis}

\centering
\footnotesize
\setlength{\tabcolsep}{1.2pt}
\renewcommand{\arraystretch}{1.10}

\begin{tabular*}{\textwidth}{
@{\extracolsep{\fill}}
lcc|cc|cc|cc|cc
@{}
}
\hline

\multirow{2}{*}{Variant / Channels}
& \multirow{2}{*}{Fuse}
& \multirow{2}{*}{Params(M)}
& \multicolumn{2}{c|}{Kvasir-SEG}
& \multicolumn{2}{c|}{DRIVE}
& \multicolumn{2}{c|}{BUSI}
& \multicolumn{2}{c}{DSB18} \\

\cline{4-11}

& &
& Dice$\uparrow$ & IoU$\uparrow$
& Dice$\uparrow$ & IoU$\uparrow$
& Dice$\uparrow$ & IoU$\uparrow$
& Dice$\uparrow$ & IoU$\uparrow$ \\

\hline

\textbf{LightMIS-T} $(3,6,10,16,24)$
& 10
& 0.017
& 87.52$\pm$1.30 & 80.96$\pm$1.45
& 81.76$\pm$1.00 & 69.26$\pm$1.28
& 79.62$\pm$1.28 & 71.48$\pm$0.96
& 91.29$\pm$0.87 & 84.89$\pm$1.20 \\

\textbf{LightMIS-S} $(5,10,16,24,40)$
& 16
& 0.038
& 88.47$\pm$1.96 & 82.37$\pm$2.27
& 81.89$\pm$0.99 & 69.48$\pm$1.25
& 80.11$\pm$1.41 & 72.31$\pm$1.03 
& 91.56$\pm$0.81 & 85.25$\pm$1.15 \\

\textit{Config. I} $(5,10,16,24,40)$
& 32 & 0.052
& 88.62$\pm$1.64 & 82.54$\pm$1.91
& 82.09$\pm$0.92 & 69.74$\pm$1.19
& \textbf{80.32$\pm$0.91} & \textbf{72.51$\pm$0.69}
& 91.52$\pm$0.83 & 85.28$\pm$1.11 \\

\textit{Config. II} $(10,20,32,48,80)$
& 16 & 0.115
& 89.17$\pm$1.78 & 83.38$\pm$2.00
& 82.02$\pm$1.07 & 69.66$\pm$1.34
& 80.12$\pm$1.44 & 72.35$\pm$1.04
& 91.54$\pm$0.81 & 85.25$\pm$1.05 \\

\textbf{LightMIS} $(10,20,32,48,80)$
& 32
& 0.131
& \textbf{89.87$\pm$1.53} & \textbf{84.25$\pm$1.70}
& \textbf{82.15$\pm$0.98} & \textbf{69.82$\pm$1.29}
& 80.16$\pm$1.23 & 72.28$\pm$1.25
& \textbf{91.81$\pm$0.67} & \textbf{85.62$\pm$0.93} \\

\hline
\end{tabular*}

\end{table*}

\begin{table}[!t]
\caption{Ablation of the third block in the proposed AFC module of LightMIS-T across six datasets. Only the final \mbox{ResAKF} block is replaced by AKF. Results are mean$\pm$standard deviation over five folds under nnU-Net~\cite{nnunet}.}

\label{tab:resakf_ablation}

\centering
\footnotesize
\setlength{\tabcolsep}{2.5pt}
\renewcommand{\arraystretch}{1.08}

\resizebox{\columnwidth}{!}{
\begin{tabular}{@{}lcc|cc@{}}
\hline

\multirow{2}{*}{Dataset}
& \multicolumn{2}{c}{Dice$\uparrow$}
& \multicolumn{2}{c}{IoU$\uparrow$} \\

\cline{2-5}

& with AKF & with ResAKF
& with AKF & with ResAKF \\

\hline

ISIC-2017
& 89.17$\pm$0.65 & 89.09$\pm$0.63
& 82.40$\pm$0.85 & 82.30$\pm$0.77 \\

ISIC-2018
& 89.87$\pm$0.27 & 89.69$\pm$0.56
& 83.42$\pm$0.35 & 83.21$\pm$0.64 \\

Kvasir-SEG
& 87.25$\pm$1.39 & 87.52$\pm$1.30
& 80.65$\pm$1.69 & 80.96$\pm$1.45 \\

DRIVE
& 81.33$\pm$1.23 & 81.76$\pm$1.00
& 68.73$\pm$1.46 & 69.26$\pm$1.28 \\

BUSI
& 79.27$\pm$1.26 & 79.62$\pm$1.28
& 71.02$\pm$1.08 & 71.48$\pm$0.96 \\

DSB18
& 91.33$\pm$0.88 & 91.29$\pm$0.87
& 84.94$\pm$1.16 & 84.89$\pm$1.20 \\

\hline

\textbf{Modality-macro}
& 85.74 & \textbf{85.92}
& 77.65 & \textbf{77.87} \\

\hline
\end{tabular}
}

\end{table}

\begin{table}[!t]
\caption{Composition ablation of the proposed AFC module in LightMIS-T on BUSI. Results are mean$\pm$standard deviation over five folds under nnU-Net~\cite{nnunet}.}

\label{tab:afc_composition}

\centering
\footnotesize
\setlength{\tabcolsep}{3.5pt}
\renewcommand{\arraystretch}{1.10}

\resizebox{\columnwidth}{!}{
\begin{tabular}{@{}lccc@{}}
\hline

Configuration
& Params (M)
& Dice$\uparrow$
& IoU$\uparrow$ \\

\hline

AKF & 0.006 & 75.17$\pm$2.35 & 66.09$\pm$2.21 \\

PRF & 0.010 & 75.33$\pm$0.94 & 66.94$\pm$0.86 \\

ResAKF & 0.006 & 75.58$\pm$1.39 & 66.54$\pm$1.32 \\

AKF $\rightarrow$ ResAKF & 0.010 & 78.14$\pm$0.88 & 69.67$\pm$0.69 \\

AKF $\rightarrow$ AKF $\rightarrow$ ResAKF & 0.014 & 78.10$\pm$1.54 & 69.83$\pm$1.55 \\

AKF $\rightarrow$ ResAKF $\rightarrow$ ResAKF & 0.014 & 78.63$\pm$1.75 & 70.40$\pm$1.36 \\

AKF $\rightarrow$ PRF & 0.014 & 79.08$\pm$1.51 & 70.90$\pm$1.41 \\

PRF $\rightarrow$ ResAKF & 0.014 & 79.20$\pm$1.17 & 71.11$\pm$0.97 \\

\textbf{AKF $\rightarrow$ PRF $\rightarrow$ ResAKF}
& 0.017
& \textbf{79.62$\pm$1.28}
& \textbf{71.48$\pm$0.96} \\

\hline
\end{tabular}
}

\end{table}

\begin{table*}[!t]
\caption{Parameter-matched comparison of prediction heads using the same LightMIS-T encoder and training protocol. Where necessary, parameter matching is achieved by adjusting the internal channel widths of the compared blocks. Dice results are reported as mean $\pm$ standard deviation over five folds under nnU-Net~\cite{nnunet}.}
\label{tab:head_ablation}

\centering
\footnotesize
\setlength{\tabcolsep}{1.2pt}
\renewcommand{\arraystretch}{1.10}

\begin{tabularx}{\linewidth}{
@{}
>{\raggedright\arraybackslash}p{0.36\linewidth}
|
>{\centering\arraybackslash}p{0.09\linewidth}
>{\centering\arraybackslash}p{0.08\linewidth}
|YYYY
@{}
}
\hline

\multirow{2}{*}{Head}
& \multirow{2}{*}{Params $\downarrow$}
& \multirow{2}{*}{GFLOPs $\downarrow$}
& \multicolumn{4}{c}{Dice (\%) $\uparrow$} \\

\cline{4-7}

& & &
BUSI
& Kvasir-SEG
& DRIVE
& Macro \\

\hline

Deepest feature with pointwise head
& 17,358
& 0.052
& 78.68$\pm$1.47
& 86.44$\pm$1.37
& 52.62$\pm$1.94
& 72.58 \\

Aligned element-wise sum
& 17,377
& 0.102
& 78.47$\pm$1.84
& 86.40$\pm$2.25
& 81.39$\pm$1.12
& 82.09 \\

Aligned concat $+\,1{\times}1$ projection
& 17,330
& 0.105
& 78.32$\pm$1.97
& 86.26$\pm$1.85
& 81.44$\pm$1.27
& 82.01 \\

FPN-lite~\cite{fpn}
& 17,387
& 0.139
& 78.77$\pm$1.37
& 87.23$\pm$1.41
& 81.69$\pm$1.21
& 82.56 \\

Parameter-matched U-Net decoder
& 17,402
& 0.093
& 78.47$\pm$1.58
& 87.00$\pm$1.65
& 81.68$\pm$0.89
& 82.38 \\

\textbf{SAP $+$ final AFC (proposed)}
& 17,400
& 0.099
& \textbf{79.62$\pm$1.28}
& \textbf{87.52$\pm$1.30}
& \textbf{81.76$\pm$1.00}
& \textbf{82.97} \\

\hline
\end{tabularx}
\end{table*}

\begin{table*}[!t]

\caption{Parameter-matched comparison of compact feature-processing blocks used to replace the PRF components within the AFC modules of the five LightMIS-T encoder stages. Where necessary, parameter matching is achieved by adjusting the internal channel widths of the compared blocks. Dice results are reported as mean$\pm$standard deviation over five folds under nnU-Net~\cite{nnunet}.}

\label{tab:prf_comparison}

\centering
\footnotesize
\setlength{\tabcolsep}{1.2pt}
\renewcommand{\arraystretch}{1.10}

\begin{tabularx}{\linewidth}{
@{}
>{\raggedright\arraybackslash}p{0.36\linewidth}
|
>{\centering\arraybackslash}p{0.09\linewidth}
>{\centering\arraybackslash}p{0.08\linewidth}
|YYYY
@{}
}
\hline

\multirow{2}{*}{Encoder block}
& \multirow{2}{*}{Params $\downarrow$}
& \multirow{2}{*}{GFLOPs $\downarrow$}
& \multicolumn{4}{c}{Dice (\%) $\uparrow$} \\

\cline{4-7}

& &
& BUSI
& Kvasir-SEG
& DRIVE
& Macro \\

\hline

Depthwise $3{\times}3$ residual block & 17,398 & 0.098 & 78.89$\pm$1.72 & \textbf{87.66$\pm$1.62} & 81.71$\pm$1.02 & 82.75 \\
Parallel multi-kernel depthwise block & 17,415 & 0.100 & 78.99$\pm$1.42 & 87.28$\pm$1.50 & 81.67$\pm$1.20 & 82.65 \\
CMRF-style cascade~\cite{tinyunet} & 17,446 & 0.116 & 78.70$\pm$1.16 & 87.51$\pm$1.76 & 81.66$\pm$0.80 & 82.62 \\
AKF~\cite{aulunet} & 17,418 & 0.096 & 78.63$\pm$2.20 & 86.83$\pm$2.22 & 81.46$\pm$1.36 & 82.31 \\
PRF without progressive transfer & 17,400 & 0.099 & 79.33$\pm$1.65 & 87.59$\pm$1.66 & 81.57$\pm$0.92 & 82.83 \\
\textbf{PRF (proposed)} & 17,400 & 0.099 & \textbf{79.62$\pm$1.28} & 87.52$\pm$1.30 & \textbf{81.76$\pm$1.00} & \textbf{82.97} \\

\hline
\end{tabularx}

\end{table*}

\subsection{Ablation Studies}
\label{sec:ablation}

We perform architectural and module-level ablation experiments to validate the major design choices, including the encoder depth, the internal PRF design, the scaling strategy, and the composition of the proposed AFC module.

The five-level configuration provides the most favorable accuracy--complexity trade-off among the tested encoder depths, as shown in Table~\ref{tab:arch_ablation}. Although the sixth level slightly improves selected Kvasir-SEG boundary metrics, it approximately doubles the parameter count and reduces the corresponding overlap scores.

The PRF ablation experiments, performed on both the encoder and the final AFC block of the aggregation head, show that the effect of each operation is dataset-dependent, as reported in Table~\ref{tab:arch_ablation}. Receptive-field diversity improves all reported metrics on BUSI and Kvasir-SEG relative to uniform \(3\times3\) depthwise processing. Progressive branch interaction provides a substantial BUSI overlap improvement, whereas its Kvasir-SEG overlap effect is small and the boundary metrics slightly favor the non-interacting variant. Channel expansion contributes more clearly on Kvasir-SEG than on BUSI.

The scaling analysis isolates the effects of encoder width and aggregation width $F$, as summarized in Table~\ref{tab:scaling_analysis}. Configuration I increases $F$, whereas Configuration II widens only the encoder. Both changes contribute, motivating their joint scaling in LightMIS, which achieves the strongest overlap results on Kvasir-SEG, DRIVE, and DSB18. On BUSI, performance appears to saturate beyond LightMIS-S, with wider configurations producing marginal variations.

Replacing the final AKF with ResAKF, as shown in Table~\ref{tab:resakf_ablation}, improves the modality-macro Dice from \(85.74\%\) to \(85.92\%\), with the largest gains on Kvasir-SEG, DRIVE, and BUSI. The effect is not uniform, as small reductions occur on the two ISIC datasets and DSB18. We therefore retain ResAKF for its aggregate trade-off rather than claiming a universal per-dataset benefit.

Within the evaluated configurations, the complete AKF--PRF--ResAKF sequence obtains the highest BUSI Dice and IoU, as indicated in Table~\ref{tab:afc_composition}. Because the alternatives differ in parameter count, these results demonstrate the performance of the complete configuration but do not isolate composition independently of model capacity. Therefore, we conduct parameter-matched comparisons at both the encoder and prediction-head levels. Table~\ref{tab:head_ablation} compares parameter-matched heads using the same LightMIS-T encoder and training protocol. All share a final pointwise convolution followed by a $1{\times}1$ class projection, with internal widths adjusted for parameter matching. Deepest feature uses only the final encoder output. Aligned sum and aligned concat use the same per-level projections and spatial alignment, differing only in whether the resulting features are fused by element-wise summation or channel-wise concatenation. FPN-lite~\cite{fpn} recursively upsamples the deeper feature, adds each lateral projection, and applies depthwise processing. Compact U-Net uses four stages of upsampling, projected-skip concatenation, and depthwise-separable fusion. The results highlight the importance of early encoder features for segmenting thin retinal structures on DRIVE, where using only the deepest feature causes a substantial performance drop. However, simple aligned summation or concatenation does not outperform deepest-feature prediction on BUSI or Kvasir-SEG, indicating the importance of effective multi-scale feature fusion and processing. Among the tested configurations, the proposed method achieved the highest Dice scores across all three datasets, with the largest margin on BUSI.

Table~\ref{tab:prf_comparison} compares parameter-matched replacements for the PRF component within the AFC modules of the five encoder stages only. The tested alternatives include a residual depthwise $3{\times}3$ block; a parallel multi-kernel block that processes the same feature tensor through independent depthwise $3\times3$, dilated $3\times3$, and $5\times5$ branches; an eight-branch CMRF cascade~\cite{tinyunet}; and an AKF block~\cite{aulunet} that adaptively combines local and dilated responses. A~PRF~variant without progressive cross-branch transfer is also evaluated. The proposed PRF achieves the highest macro Dice with leading scores on BUSI and DRIVE, whereas the residual depthwise $3\times3$ block is marginally better on Kvasir-SEG. The complete configuration provides the strongest task-aggregated trade-off among the tested variants, although the contribution of individual components is dataset-dependent.

\section{Limitations}
\label{sec:limitations}

This study is restricted to public 2D binary segmentation
datasets and does not establish performance on multiclass
or volumetric segmentation. The common nnU-Net training
protocol provides a controlled architectural comparison but
may not be optimal for every baseline, particularly models
designed around pretraining, different optimizers, deep
supervision, or model-specific training schedules.

For DRIVE, ISIC-2017, and ISIC-2018, publicly labelled
official partitions were pooled before cross-validation.
Consequently, the reported results are pooled-label
cross-validation results rather than official challenge-test
results. Cross-validation was performed at image level
without subject grouping or category-based stratification.
Where multiple images may originate from the same subject,
lesion, or acquisition sequence, independence across folds
cannot be guaranteed.

Except for DRIVE, images were directly resized to
\(256\times256\) without preserving aspect ratio. This
standardizes computation but may alter anatomical or lesion
geometry. The effect of aspect-ratio-preserving resizing
remains to be evaluated.

The modality-macro average first combines ISIC-2017 and
ISIC-2018 into a single dermoscopy score and then assigns
equal weight to the five imaging modalities. Differences
between the leading models are small and should be
interpreted using paired confidence intervals. The current
paired analyses use fixed out-of-fold predictions and do not
capture variation caused by repeated model training.

The mobile evaluation covers one smartphone GPU and one
software stack. It measures delegated model execution
rather than complete application latency and does not
establish performance across other edge processors,
sustained thermal conditions, energy budgets, or clinical
workflows. No prospective clinical evaluation or external
institutional validation was performed.

\section*{Data and Code Availability}

The implementation, environment specification, exact five-fold partitions, nnU-Net plan files, dataset-conversion scripts, baseline commit hashes, training commands, checkpoints, metric implementation, FLOP-counting script, LiteRT conversion code, and desktop  and mobile benchmarking scripts are publicly available at \url{https://github.com/AndreiiArhire/LightMIS}.

The original medical-image datasets remain available from their respective providers under the terms cited in Table~\ref{tab:dataset_protocol}.

\section{Conclusion}
\label{sec:conclusion}

We introduced LightMIS, a scalable family of
ultra-lightweight direct-aggregation networks for 2D binary
medical image segmentation without a learned stage-wise
decoder. Scale-Aligned Projection blocks align all five
encoder levels to a common resolution, while the Adaptive
Fusion Cascade combines Adaptive Kernel Fusion, the
proposed Progressive Receptive Fusion module, and residual
adaptive-kernel refinement.

Under a common nnU-Net protocol across six datasets,
LightMIS achieved modality-macro Dice and IoU scores of
\(86.71\%\) and \(78.99\%\) with \(0.131\,\mathrm{M}\)
parameters and \(0.575\,\mathrm{GFLOPs}\). Mobile U-ViT
obtained the highest observed modality-macro values of
\(86.75\%\) and \(79.07\%\); therefore, the results are
described as numerically close rather than as evidence of
equivalence or statistical superiority. LightMIS nevertheless
obtained the best mean Dice rank and substantially reduced
parameter count and computational cost relative to the
strongest larger baselines.

On an Arm Mali-G52 MC2 GPU, all three LightMIS variants
achieved complete GPU delegation. Median delegated latency
ranged from \(53.31\,\mathrm{ms}\) for LightMIS-T to
\(138.31\,\mathrm{ms}\) for LightMIS. These measurements
demonstrate on-device execution feasibility for the
evaluated input size, but they do not constitute evidence of
end-to-end clinical deployment.

Future work will address patient-grouped and external
validation, multiclass and volumetric segmentation,
quantized execution on additional edge processors, and the general applicability of PRF and AFC to other dense
prediction tasks.
\section{Acknowledgments}
\label{sec:acknowledgments}
This research was conducted within the project "\textit{ArtCADe: Dezvoltarea unei platforme suport decizie pentru diagnosticul arteriopatiilor, bazată pe tehnologii Big Data, Învățare Automată și Procesare de Imagini} - Cod SMIS 338317" under the program "PR/NE/2024/P1/RSO1.1-RSO1.3/1 - PROIECTE DE CDI ȘI INVESTIȚII ÎN IMM".
\\Radu Timofte acknowledges the support of the Alexander von Humboldt Foundation.

{
    \small
    \bibliographystyle{ieeenat_fullname}
    \bibliography{main}
}


\end{document}